\documentclass[10pt,twocolumn,letterpaper]{article}

\usepackage[pagenumbers]{iccv} 
\usepackage{ulem}
\usepackage{colortbl}  
\usepackage{arydshln}
\usepackage{bm}
\usepackage[pagenumbers]{iccv} 

\usepackage{multirow}
\usepackage{graphicx}

\definecolor{iccvblue}{rgb}{0.21,0.49,0.74}
\usepackage[pagebackref,breaklinks,colorlinks,allcolors=iccvblue]{hyperref}

\def\paperID{1139} 
\def\confName{ICCV}
\def\confYear{2025}

\title{Towards Synthesized and Editable Motion In-Betweening Through Part-Wise Phase Representation}

\author{
    Minyue Dai\textsuperscript{\rm 1,\rm 2} \quad
    Ke Fan\textsuperscript{\rm 3}\quad
    Bin Ji\textsuperscript{\rm 3}\quad
    Haoran Xu\textsuperscript{\rm 4}\quad
    Haoyu Zhao\textsuperscript{\rm 5}\\
    Junting Dong\textsuperscript{\rm 2}\quad
    Jingbo Wang\textsuperscript{\rm 2}\quad
    Bo Dai\textsuperscript{\rm 6}\\
    \textsuperscript{1}Fudan University \quad
    \textsuperscript{2}Shanghai Artificial Intelligence Laboratory \quad 
    \textsuperscript{3}Shanghai Jiaotong University\\
    \textsuperscript{4}Zhejiang University \quad 
    \textsuperscript{5}Wuhan University \quad 
    \textsuperscript{6}University of Hong Kong \quad 
}

\begin{document}

\twocolumn[{
\renewcommand\twocolumn[1][]{#1}
\maketitle
\begin{center}
    \centering
    \captionsetup{type=figure}
    \includegraphics[width=1\textwidth]{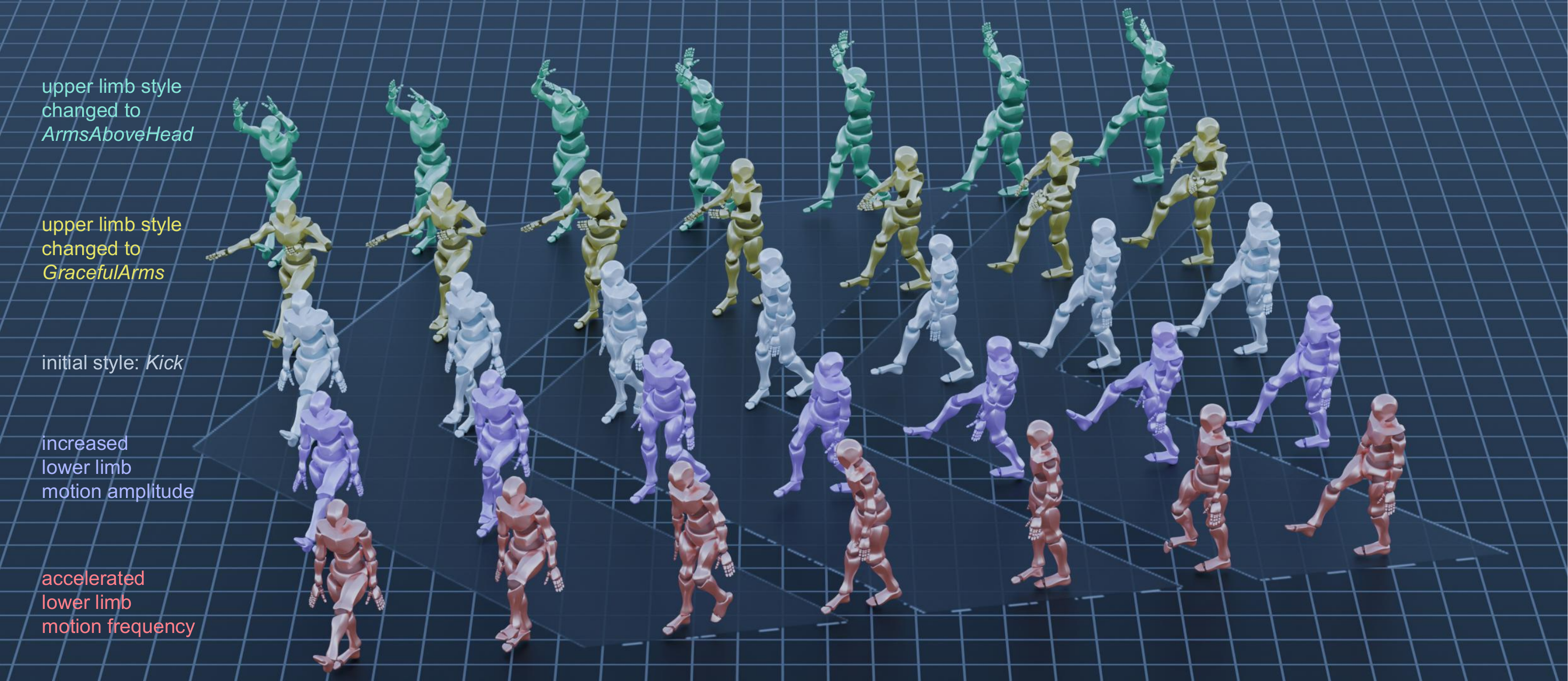}
    \vspace{-15pt}
    \captionof{figure}{We demonstrate the effectiveness of our framework in generating stylized online motion in-between. It excels at producing realistic animations that accurately reflect the target style during variations in body part movements, while also allowing adjustments to the individual body parts or overall movements. These capabilities make our framework a robust and versatile solution for stylized in-between generation through part-aware phase representation.}
    \label{fig:teaser}
\end{center}
}]

\maketitle
\begin{abstract}
Styled motion in-betweening is crucial for computer animation and gaming. However, existing methods typically encode motion styles by modeling whole-body motions, often overlooking the representation of individual body parts. This limitation reduces the flexibility of infilled motion, particularly in adjusting the motion styles of specific limbs independently. To overcome this challenge, we propose a novel framework that models motion styles at the body-part level, enhancing both the diversity and controllability of infilled motions. Our approach enables more nuanced and expressive animations by allowing precise modifications to individual limb motions while maintaining overall motion coherence. Leveraging phase-related insights, our framework employs periodic autoencoders to automatically extract the phase of each body part, capturing distinctive local style features. Additionally, we effectively decouple the motion source from synthesis control by integrating motion manifold learning and conditional generation techniques from both image and motion domains. This allows the motion source to generate high-quality motions across various styles, with extracted motion and style features readily available for controlled synthesis in subsequent tasks. Comprehensive evaluations demonstrate that our method achieves superior speed, robust generalization, and effective generation of extended motion sequences.
\end{abstract}
\section{Introduction}
\label{sec:Introduction}
Motion in-betweening is a fundamental technique for generating complete motion sequences from sparse keyframe poses. It serves as a cornerstone for numerous applications in computer vision and computer graphics, including character animation~\cite{zhao2024sg,zhao2024chase}, action recognition, and motion capture. The diversity and controllability of in-between motion are of paramount importance, as they directly influence the realism, expressiveness, and adaptability of synthesized motion. In this work, we introduce a novel approach to fine-grained style control in motion in-betweening, explicitly focusing on the independent manipulation of motion styles across different body parts.

Early methods formulated in-between motions as a motion planning problem~\cite{wang2014energy,wang2013harmonic,ye2010synthesis}. In the context of deep learning, in-between motion is framed as a task within motion manifold learning~\cite{chen2020dynamic,holden2016deep,wang2019spatio,tang2022CVAE} or as a control problem using temporal control signals~\cite{ling2020character}. However, without precise style control, character motion transitions generated by these methods often default to the most probable motion, misaligning with the intended style. Although some recent approaches encode style information from entire motion sequences~\cite{tang2023RSMT}, they inadequately capture body part style features, particularly during stylistic changes. This limitation results in generating the most probable motion rather than the desired stylistic expression when adjusting motion of body parts. Furthermore, these methods do not allow for adjusting of body part motions.

To address this issue, we propose a novel framework for stylized online motion in-betweening that significantly enhances the diversity and controllability of in-between motion styles. Our method draws on phase-related techniques, utilizing periodic autoencoders~\cite{starke2022deepphase} to automatically extract parameters such as frequency amplitudes and motion velocities within the motion latent space, ensuring high-quality in-between motion. However, existing approaches typically extract phase information from the entire body, limiting precise style control over specific body parts during in-between motion, such as maintaining lower body motion while altering upper body dynamics.

Therefore, our framework first extracts local phases of body parts using the character's kinematic chain, enabling individual modeling of the motion latent space for each limb. Leveraging this pre-captured latent space, we propose our framework to infill motion at the subsequent time step with an additional motion sampler. To better align the motion style with part-level motions, we encode the motion style through body part phases, effectively capturing specific stylistic attributes without relying on the whole body representation~\cite{mason2022real, tang2023RSMT}. This approach accurately reflects the target style during changes in body part motions, such as raising legs while keeping hands behind the back, thereby unlocking the potential for fine-grained style control in the in-between motion. Consequently, our method demonstrates superior generalization capabilities. The fine-grained phase representation facilitates nuanced adjustments to the amplitude of individual body parts or overall motion. We can modify motion characteristics by scaling the amplitude and frequency of the body part phases while preserving overall body coordination. Extensive experiments on the 100STYLE dataset~\cite{mason2022real} demonstrate that our method can \textit{controls motion of body part motion}, ensuring consistent style adherence in long sequences and style transitions. In summary, our work makes the following contributions:
\begin{itemize}
    \item We introduce body part-based phase features, enhancing motion generation quality for long sequences.
    \item We propose a style encoder based on body part phases that captures the stylistic information of body parts.
    \item Our model allows for the adjustment of both the magnitude and frequency of body part motions.
\end{itemize}


\section{Related Works}
\paragraph{In-between motion synthesis.}
In-between motion synthesis can be formulated as a motion planning problem~\cite{arikan2002interactive,beaudoin2008motion,levine2012continuous,safonova2007construction,wang2013harmonic}, which involves solving complex optimization problems under various constraints. Data-driven methods improve efficiency and enhance motion naturalness by searching structured data such as motion graphs~\cite{kovar2023motion,min2012motion,shen2017posture}. 
As the field of neural networks continues to advance, some methodologies have emerged that focus on learning the motion manifold~\cite{holden2015learning,wang2019spatio,mo2023continuous,jang2020constructing,chen2020dynamic,he2022nemf,li2021task,petrovich2021action,rempe2021humor} to synthesize motions using various control signals~\cite{harvey2018recurrent}. Harvey et al.~\cite{harvey2020robust} proposed a neural network capable of generating plausible intermediate motions between given keyframe poses. Their primary contribution lies in strengthening the relationship between motion and time by embedding temporal information. Building upon this, Tang et al.~\cite{tang2022CVAE} introduced Convolutional Variational Autoencoder (CVAE) to further explore the impact of temporal information. 
SKEL-Betweener~\cite{agrawal2024skel} generates extended motion sequences from two poses using neural motion curves (intuitive joint-level controls for positioning and orientation). 
Our phase-based control approach complements trajectory-level methods: whereas prior work~\cite{agrawal2024skel} ensures precision through spatiotemporal curves, our method maintains natural coordination by leveraging biomechanically constrained phase parameters. This enables style-consistent editing (e.g., adjusting arm swing magnitude throughout dance sequences) without requiring manual coordination efforts.

\begin{figure*}[tphb]  
    \centering
    \includegraphics[width=0.85\textwidth]{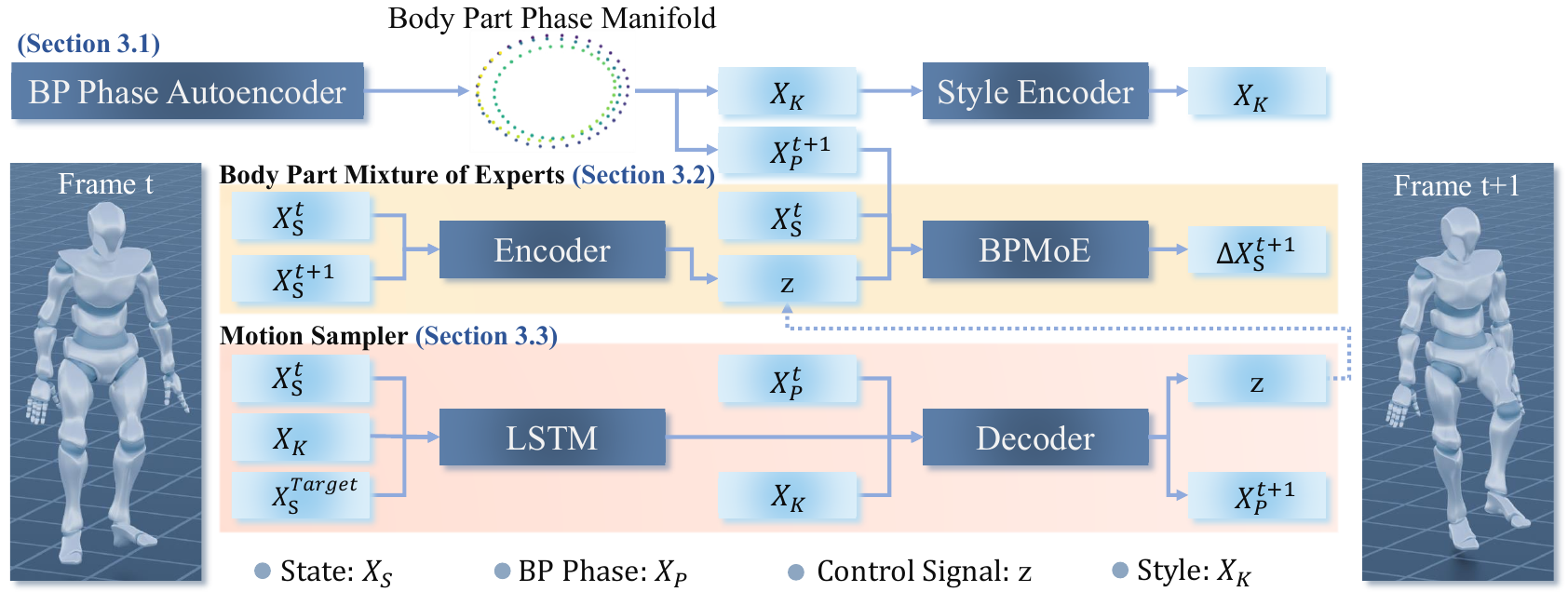}
    \caption{\textbf{System Overview.}
We first train the BP Phase Autoencoder (Body Part Phase Autoencoder) similar to \cite{starke2022deepphase}. 
Next, we train the BPMoE (Body Part Mixture of Experts). The encoder takes the current state and the next state to generate the control signal. BPMoE takes the control signal, the current state, and the next BP Phase to predict the next state.
Finally, when training the Motion Sampler, we remove the encoder, fix BPMoE and connect the Motion Sampler to BPMoE. The style encoder encodes the BP Phase into style information. The LSTM network takes the current, target state, and style information as input, and the output is decoded into control signals and the next BP Phase, along with the current BP Phase and style information.}
    \label{fig:motion sampler}
\end{figure*}

\vspace{-10pt}
\paragraph{Phase.}
Previous motion interpolation techniques struggled to adapt to longer durations between keyframes due to the generalization effects of temporal information. The first successful exploration of this issue was presented in the Phase-Functioned Neural Networks (PFNN)~\cite{holden2017phase}. It utilizes phase-controlled neural motion controllers to interpolate human motion behaviors. Based on this approach, Starke et al.~\cite{starke2019neural} proposed representing different motion skills through multiple phase variables, enabling the model to effectively learn a vast array of motions. 
Subsequently, Starke et al.~\cite{starke2020local} delved into phase representation, introducing a contact-based method for phase extraction at the local joint level. However, this approach is limited to movements involving physical contacts. Many motions, especially those of the hands, lack such contacts, making phase definition a challenging problem. Another challenge with phases based on contacts is that they require careful tuning of thresholds for computing the phase labels.
Mason et al.~\cite{mason2022real} use a PCA heuristic to compute the local phase of cyclic arm movements during locomotion, although a different heuristic would be needed for arbitrary movements.
Recently, research has shifted towards learning-based phase extraction to achieve spatiotemporal alignment for arbitrary character motions~\cite{starke2022deepphase}. This method learns multi-dimensional phase variables from arbitrary unstructured motion capture data. However, it extracts phase information from the entire body, which limits precise style control over individual body parts.
We further investigate the impact of learning-based phase extraction at the body-part level on motion quality, as well as the effects of body-part phase alterations on motion generation. Additionally, we analyze the relationship between phase and motion style in greater depth.

\section{Methodology}
To overcome the limitations of existing methods that encode motion styles holistically and lack precise part-level control\cite{tang2023RSMT,tang2022CVAE,starke2023motion,harvey2020robust}, our framework introduces a body-part decomposition strategy inspired by phase dynamics in human motion\cite{starke2022deepphase}.
Our framework comprises three distinct components:~the \textbf{BP Phase Autoencoder}~(Body Part Phase Autoencoder), the \textbf{BPMoE} (Body Part Mixture of Experts) and the \textbf{Motion Sampler}.  
First, we present the BP Phase Autoencoder, which extracts body-part phases from motion data~(Section\ref{sec:Body Part Phase Autoencoder}).
Next, the BPMoE generates the next frame's state based on the current frame's state and the control signal~(Section\ref{sec:Body Part Mixture of Experts}).
Finally, the Motion Sampler produces a control signal that aligns with the target frame, the desired transition duration, and style information~(Section\ref{sec:Motion Sampler}).
\subsection{Body Part Phase Autoencoder}
\label{sec:Body Part Phase Autoencoder}
Motion in-betweening fundamentally relies on temporally coherent transitions\cite{tang2023RSMT,tang2022CVAE,starke2023motion,harvey2020robust,zhou2020generative}. The phase variable intrinsically encodes motion dynamics by modeling cyclic patterns\cite{starke2022deepphase,holden2017phase,starke2020local}. However, conventional implementations rely on whole-body monolithic representations for phase extraction\cite{tang2023RSMT}, leading to phase entanglement, where distinct limb kinematics are coupled in latent space. This architectural limitation fundamentally restricts fine-grained control over individual body parts during motion synthesis. 
We propose part-specific temporal autoencoders that map body motions to localized phase manifolds.
The BP Phase Autoencoder takes a sequence of motion data within a specified time window \(\textbf{M} = \{X_S^0,\ldots,X_S^T\}\) as input and outputs the phases~\(\bm{X_{\mathcal{P}}^t} = \{\Theta^t_1, \Theta^t_2, \ldots, \Theta^t_n \}\).
\(\bm{X^i_S} = \{p^i, v^i, r^i\}\) includes bone positions \(p^i \in \mathbb{R}^{3B}\), bone rotations \(r^i \in \mathbb{R}^{6B}\), and bone velocities \(v^i \in \mathbb{R}^{3B}\), where \(B\) represents the number of joints.
\(\Theta_i \in \mathbb{R}^{2m} \) represents the phase information for the \(i\)th body part. \(n\) and \(m\) represent the number of body parts and the number of channels for body part phases, respectively.

We structure the latent space with explicit periodic constraints. Each body part's phase vector \(\Theta_i \) is parameterized by:
\begin{align}
\label{eq:phase}
    \Theta _i^{2j-1} = A_i^j \cdot \sin(2\pi \cdot S_i^j), 
    \Theta _i^{2j} = A_i^j \cdot \cos(2\pi \cdot S_i^j),
\end{align}
where \(A_i^j \in \mathbb{R}^{+} \) represents the magnitude and could control the motion amplitude, and \(S_i^j \in [0,1)\) as the angular progression, could regulate timing for the \(j\)th phase channel of the \(i\)th body part.

\subsection{Body Part Mixture of Experts}
\label{sec:Body Part Mixture of Experts}
Traditional motion generation systems typically rely on a single network to handle all body parts simultaneously\cite{harvey2020robust}, potentially leading to conflicting control signals. Our BPMoE solves this problem by employing specialized subnetworks for distinct biomechanical patterns, dynamically fused through phase-synchronized gating. It preserves localized motion fidelity while maintaining global coordination.
The BPMoE predicts the incremental change between frames based on the posture, control signal, and body part phases at the current frame t, which is formulated as:
\begin{align}
    &X_S^{t+1} = X_S^t + BPMoE(z^{t+1}, X_S^t, X_{\mathcal{P}}^t),
\end{align}
where the control signal \(\bm{z} \in \mathbb{R}^{32}\) governs the system's dynamics, and \(\bm{X_K} \in \mathbb{R}^{111\times512}\) captures style attributes.

During training, the encoder transforms two successive frames \(X_S^{t}, X_S^{t+1}\) into \(z^t\), which is restricted within the range \(z \sim N(0, 1)\) to capture the stochasticity of transition.
The gating network establishes a weighting mechanism for experts based on the provided input. It considers the phases extracted from the body part phase manifold and the control signal to calculate blending coefficients for the experts. The final output is a weighted combination of outputs generated by all participating experts.

\subsection{Motion Sampler}
\label{sec:Motion Sampler}
The Motion Sampler serves as a parametric controller that resolves spatio-temporal constraints. It receives inputs including the current frame \(X_S^t\), the target frame \(X_S^{Target}\), the phase of the current body part \(X_{\mathcal{P}}^t\) and the target style \(X_K\). Subsequently, it generates the control signal \(z^t\), which can be further used in the BPMoE~(Section\ref{sec:Body Part Mixture of Experts}) to sample the following frame.

Given that the phase effectively captures stylistic information, we utilize the body part phases as input for the style encoder. We first extract a style code from the body part phases using the style encoder based on convolutional neural networks: \(X_K = E_{style}(\Theta_1, \Theta_2, \ldots, \Theta_n).\)
By separating and encoding the comprehensive body motion sequence into individual body part motion phases, we can discern styles for each body part. 
Then, we use a long-short-term memory (LSTM) network to derive the next state, which is then integrated into the decoder along with the phase \(X_{\mathcal{P}}^t\) and the style code \(X_K\). This process yields the control signal \(z^t\), which is subsequently utilized within the BPMoE framework to sample the next frame.

For every body part, we anticipate an intermediate subsequent phase \(\hat{\Theta}^{t+1}\), along with estimating the amplitude \(\hat{A}^{t+1}\) and frequency \(\hat{F}^{t+1}\). An additional intermediate phase vector is calculated as follows: \(\tilde{\Theta}^{t+1} = \hat{A}^{t+1} \cdot (R(2\pi \Delta t \hat{F}^{t+1}) \cdot \Theta^t)\), where \(\Delta t\) denotes the time step, and \(R\) represents a \(2D\) rotation matrix. We apply spherical linear interpolation with a weight of 0.5 to interpolate the angles of \(\tilde{\Theta}^{t+1}\) and \(\hat{\Theta}^{t+1}\), followed by linear interpolation with the same weight for the final prediction \(\Theta^{t+1}\).
\subsection{Training.}
\label{sec:training}
Our framework employs a two-stage training strategy to progressively optimize motion quality.The first stage trains the BPMoE, followed by the training of the motion sampler in the second stage. When training the Motion Sampler, we remove the encoder, fix BPMoE and connect the Motion Sampler to BPMoE.

The first stage trains \textbf{the BPMoE} using three fundamental components.
\textbf{Reconstruction Loss} ($L_{\text{rec}}$): The L2 distance $\|X_S - X'_S\|_2^2$ between predicted ($X'_S$) and ground truth ($X_S$) joint parameters enforces precise pose reconstruction, preserving local limb geometries and global body coordination. \textbf{Latent Regularization} ($L_{\text{kl}}$): The KL-divergence term $-\frac{1}{2}(1 + \sigma^2 - \mu^2 - e^\sigma)$ regularizes the control signal distribution $z \sim \mathcal{N}(0,1)$, preventing overfitting while maintaining a smooth latent space for diverse motion sampling. The hyperparameter $\beta=0.001$ balances reconstruction accuracy and generative diversity \cite{tang2022CVAE}. \textbf{Foot Plant Constraint} ($L_{\text{foot}}$): To eliminate foot sliding, we introduce $\|f'_v \odot \delta(v_f)\|_2^2$ where $\delta(v_f)$ implements velocity-dependent contact weighting.
The composite BPMoE objective integrates these components:
\begin{equation}
\mathcal{L}_{\text{BPMoE}} = L_{\text{rec}} + \beta L_{\text{kl}} + L_{\text{foot}}.
\end{equation}
For \textbf{the motion sampler}, we extend the loss formulation with temporal and phase-aware terms:
\textbf{Temporal Consistency} ($L_{\text{rec}}$): L1 distance $\|X_S - X'_S\|_1$ ensures robust frame-to-frame coherence under motion variations.
\textbf{Keyframe Alignment} ($L_{\text{last}}$): Strict terminal constraint $\|X_S^T - (X_S^T)'\|_1$ enforces precise matching of the final generated frame to the target pose.
\textbf{Phase Dynamics} ($L_{\text{phase}}$): Maintains limb-specific motion characteristics through:
\begin{equation}
    \|A-\hat{A}\|_2^2 + \|F-\hat{F}\|_2^2 + \frac{1}{2}(\|p-\hat{p}\|_2^2 + \|p-\tilde{p}\|_2^2)
\end{equation}
, where phase velocity $F^t$ preserves cyclic continuity via:
$
F^t = (S^t - S^{t-1}) - \lfloor S^t - S^{t-1} + 0.5 \rfloor
$
.The sampler's total loss combines these objectives:
\begin{equation}
    \mathcal{L}_{\text{Sampler}} = L_{\text{rec}} + L_{\text{last}} + \lambda_{\text{foot}}L_{\text{foot}} + \lambda_{\text{phase}}L_{\text{phase}}.
\end{equation}
\begin{figure}[htbp]
    \centering
    \includegraphics[width=0.4\textwidth]{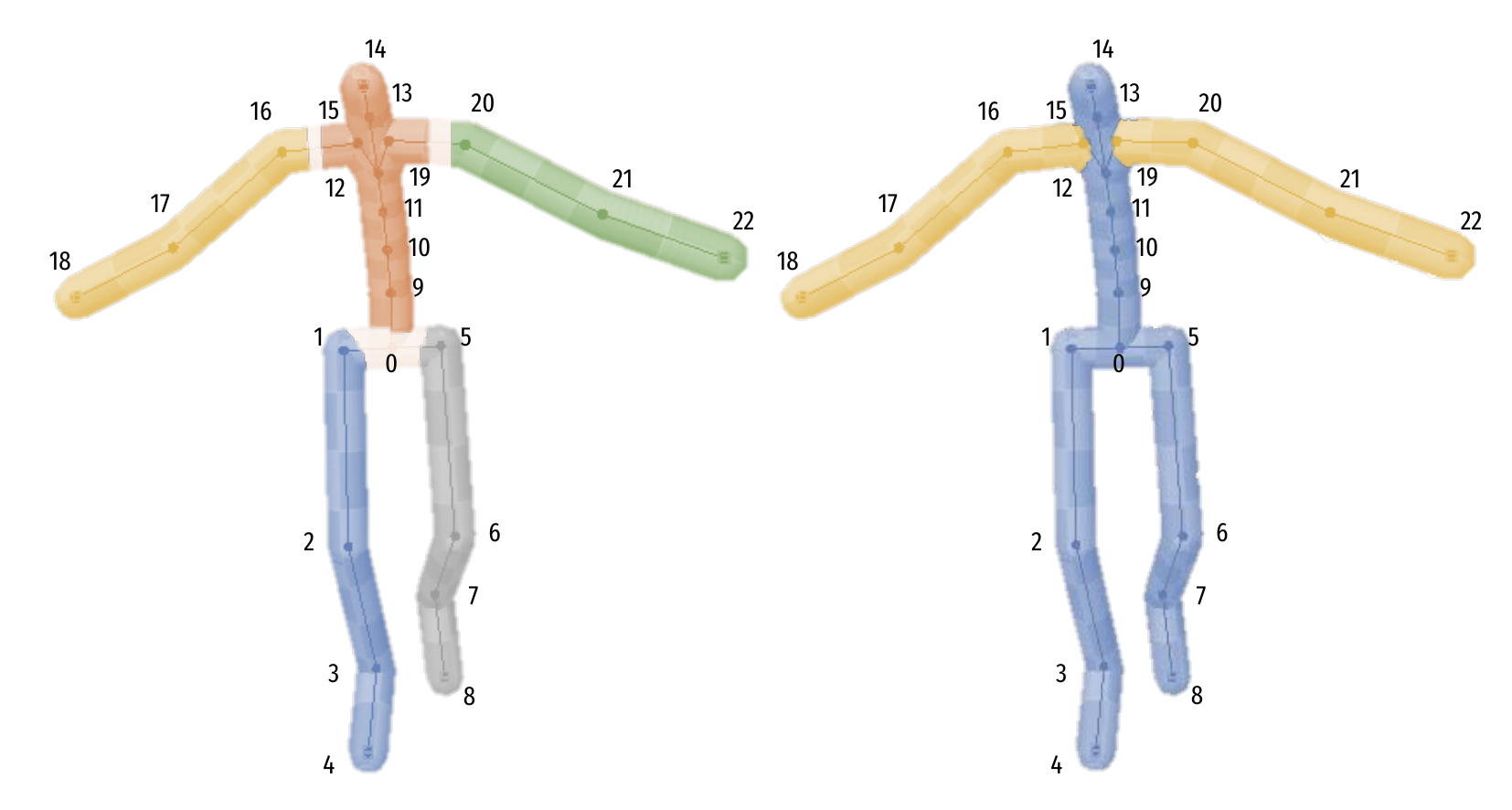}
    \captionsetup{aboveskip=1pt, belowskip=0pt}  
    \caption{The five body parts of the first method consist of the left upper limb (joints 15 to 18), right upper limb (joints 19 to 22), left lower limb (joints 1 to 4), right lower limb (joints 5 to 8), and the torso (joints 9 to 12). The second method requires the selection of two body parts: the upper limbs (joints 15 to 22) and the other segments (joints 0 to 14).}
    \label{fig:body part}
\end{figure}
\begin{table*}[thbp]
    \centering
    \small
    {
    \begin{tabular}{l|ccc|ccc|ccc}
    \toprule
        Metrics  & \multicolumn{3}{c|}{NPSS(\(\downarrow\))} & \multicolumn{3}{c|}{L2 norm of global position(\(\downarrow\))} & \multicolumn{3}{c}{skating(\(\downarrow\))} \\ \midrule
        Frames  
            & 120  & 140  & 160 
            & 120  & 140  & 160 
            & 120  & 140  & 160\\
      \midrule
      \rowcolor{gray!10}CVAE
            &  7.179  & 10.788 & 14.645 
            & 27.936 & 34.322 & 39.731 
            &  0.127 &  0.111 &  0.097\\
      RSMT             
            &  7.378  & 10.311 & 13.981 
            & 25.848 & 32.082 & 36.664 
            &  0.137 &  0.129 &  0.116\\
      \rowcolor{gray!10}PhaseMIB
            &7.133 &10.512 &12.171
            &25.768 &31.586 &35.798
            &0.142 &0.132 &0.126  \\
      OURS(2)           
            &  \textbf{6.780}  &  \textbf{9.443} & \textbf{12.178} 
            & \textbf{25.300} & \textbf{30.052} & \textbf{35.185} 
            &  0.116 &  0.100 &  0.093\\
      \rowcolor{gray!10}OURS(5)           
            &  7.800  & 10.909 & 14.626 
            & 25.816 & 31.118 & 35.746 
            &  \textbf{0.094} &  \textbf{0.083} &  \textbf{0.074}\\
    \bottomrule
    \end{tabular}}
    \caption{Comparison on reconstruction and foot skating metrics of different methods.}
    \label{tab:Comparison A}
\end{table*}

\section{Experiments}
In this section, we first describe our experimental setup and evaluation metrics. Next, we analyze the effectiveness of the proposed framework. Finally, we showcase our method's capability for parametric control of limb motion characteristics through amplitude-frequency modulation in synthesized animations.
\begin{table*}[t!]
\centering

\small
{
\begin{tabular}{l|ccc|ccc|ccc}
\toprule
NPSS (\(\downarrow\))  & \multicolumn{3}{c|}{upper limbs} & \multicolumn{3}{c|}{left upper limb} & \multicolumn{3}{c}{right upper limb} \\ \midrule
Frames  
                & 120  & 140  & 160 
                & 120  & 140  & 160 
                & 120  & 140  & 160\\ \midrule
\rowcolor{gray!10}CVAE           
                &  7.279  & 10.968  & 14.821        
                &  7.289  & 10.852  & 14.763        
                &  7.253  & 10.901  & 14.817    \\
RSMT(original)
                &  7.266  & 10.525  & 14.113        
                &  7.312  & 10.567  & 14.038        
                &  7.384  & 10.453  & 14.178    \\
\rowcolor{gray!10}RSMT(modified)
                &  7.297  & 10.426  & 13.909        
                &  7.221  & 10.466  & 13.918        
                &  7.372  & 10.521  & 14.197    \\
PhaseMIB 
                &  7.158  &  10.524  & 12.985        
                &  7.667  &  10.686  &  12.769
                &  7.731  &  10.705 & 12.810
                    \\
\rowcolor{gray!10}OURS(2)         
                &  \textbf{6.938}  &  \textbf{9.633}  & \textbf{12.364}        
                &  \textbf{6.928}  &  \textbf{9.603}  & \textbf{12.333}        
                &  \textbf{6.885}  &  \textbf{9.634}  & \textbf{12.307}    \\
OURS(5)         
                &  7.779  & 10.943  & 14.710        
                &  7.795  & 10.950  & 14.455        
                &  7.913  & 11.075  & 14.672    \\



\toprule

L2 norm (\(\downarrow\))  
& \multicolumn{3}{c|}{upper limbs} & \multicolumn{3}{c|}{left upper limb} & \multicolumn{3}{c}{right upper limb} \\ \midrule
                & 120  & 140  & 160 
                & 120  & 140  & 160 
                & 120  & 140  & 160\\ \midrule
\rowcolor{gray!10}CVAE           
                & 28.343  & 34.586  & 40.027        
                & 28.220  & 34.440  & 40.036        
                & 28.256  & 34.621  & 39.970    \\
RSMT(original)
                & 26.314  & 32.333  & 36.925        
                & 26.087  & 32.196  & 36.899        
                & 26.233  & 32.297  & 37.042    \\
\rowcolor{gray!10}RSMT(modified)
                & 26.255  & 32.303  & 36.998        
                & 26.074  & 32.164  & 36.790        
                & 26.185  & 32.219  & 37.033    \\
PhaseMIB 
                &  27.124  &  31.714 & 35.946
                &  26.537  &  33.322  &  36.754
                &  27.934  &  34.386 & 36.967
                \\
\rowcolor{gray!10}OURS(2)         
                & \textbf{26.027}  & \textbf{30.781}  & \textbf{35.701}        
                & \textbf{25.798}  & \textbf{30.518}  & \textbf{35.387}        
                & \textbf{25.675}  & \textbf{30.451}  & \textbf{35.454}    \\
OURS(5)         
                & 26.113  & 31.543  & 36.242        
                & 25.995  & 31.424  & 36.029        
                & 26.162  & 31.342  & 35.969    \\


\toprule

Skating (\(\downarrow\))  & \multicolumn{3}{c|}{upper limbs} & \multicolumn{3}{c|}{left upper limb} & \multicolumn{3}{c}{right upper limb} \\ \midrule
                & 120  & 140  & 160 
                & 120  & 140  & 160 
                & 120  & 140  & 160\\ \midrule
GT              
                & 0.044  &  0.044   &  0.044        
                & 0.044     &  0.044  &  0.044      
                & 0.044   &  0.044  &  0.044     \\
\hdashline
\rowcolor{gray!10}CVAE           
                & 0.132   &  0.114  &  0.100        
                &  0.130    &  0.112  &  0.099      
                &  0.130  &  0.113  &  0.100    \\
RSMT(original)
                & 0.141   &  0.134  &  0.120        
                &  0.141    &  0.134  &  0.122      
                &  0.138  &  0.130  &  0.117    \\
\rowcolor{gray!10}RSMT(modified)
                & 0.141   &  0.131  &  0.119        
                &  0.142    &  0.130  &  0.120      
                &  0.138  &  0.129  &  0.118    \\
PhaseMIB 
                & 0.146 & 0.139 &0.126    
                & 0.152 & 0.137 & 0.130 
                & 0.154 & 0.143 & 0.137
                \\
\rowcolor{gray!10}OURS(2)         
                & 0.115   &  0.105  &  0.094        
                &  0.115    &  0.102  &  0.091      
                &  0.116  &  0.105  &  0.093    \\
OURS(5)         
                & \textbf{0.099}   &  \textbf{0.085}  &  \textbf{0.078}        
                &  \textbf{0.095}    &  \textbf{0.083}  &  \textbf{0.076}      
                &  \textbf{0.099}  &  \textbf{0.086}  &  \textbf{0.077}    \\

\bottomrule
\end{tabular}}

\caption{Comparison of reconstruction and foot skating metrics for different methods with changes in body part styles}
\label{tab:compare in AB}
\end{table*}
\begin{table*}[htbp]
    \centering

    \small
    {
    \begin{tabular}{l|ccc|ccc|ccc}
    \toprule
        Metrics  & \multicolumn{3}{c|}{NPSS(\(\downarrow\))} & \multicolumn{3}{c|}{L2 norm of global position(\(\downarrow\))} & \multicolumn{3}{c}{skating(\(\downarrow\))} \\ \midrule
        Frames  
                & 120  & 140  & 160 
                & 120  & 140  & 160 
                & 120  & 140  & 160\\
      \midrule
      \rowcolor{gray!10}Motion-based      
                &  7.573  & 10.927 & 14.616 
                & 27.551 & 32.609 & 37.184 
                &  0.148 &  0.126 &  0.109 \\
      OURS(2)           
                &  \textbf{6.780}  &  \textbf{9.443} & \textbf{12.178} 
                & \textbf{25.300} & \textbf{30.052} & \textbf{35.185} 
                & 0.116  &  0.100 &  0.093 \\
      \rowcolor{gray!10}OURS(5)          
                &  7.800  & 10.909 & 14.626 
                & 25.816 & 31.118 & 35.746 
                &  \textbf{0.094} &  \textbf{0.083} &  \textbf{0.074}\\
        
    \bottomrule
    \end{tabular}}

    \caption{Comparisons of reconstruction and foot skating between motion-based and body-part-phase-based representation.}
    \label{tab:style A}
\end{table*}
\begin{figure*}[thbp]
    \centering
    \begin{subfigure}{0.275\textwidth}
        \centering
        \includegraphics[width=\linewidth]{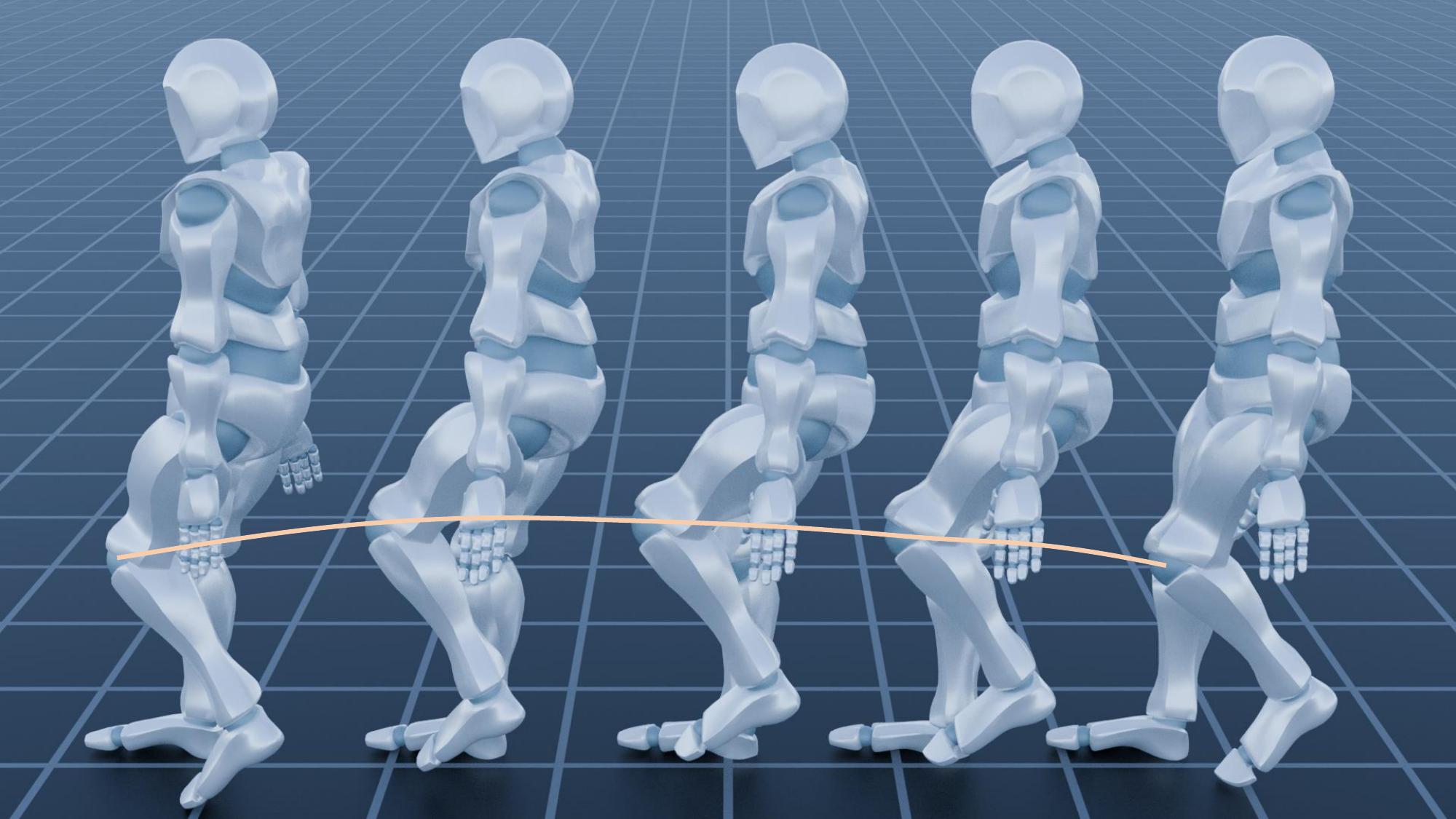}
        \caption{}
        \label{fig:change A a}
    \end{subfigure}%
    \hspace{20px} 
    \begin{subfigure}{0.275\textwidth}
        \centering
        \includegraphics[width=\linewidth]{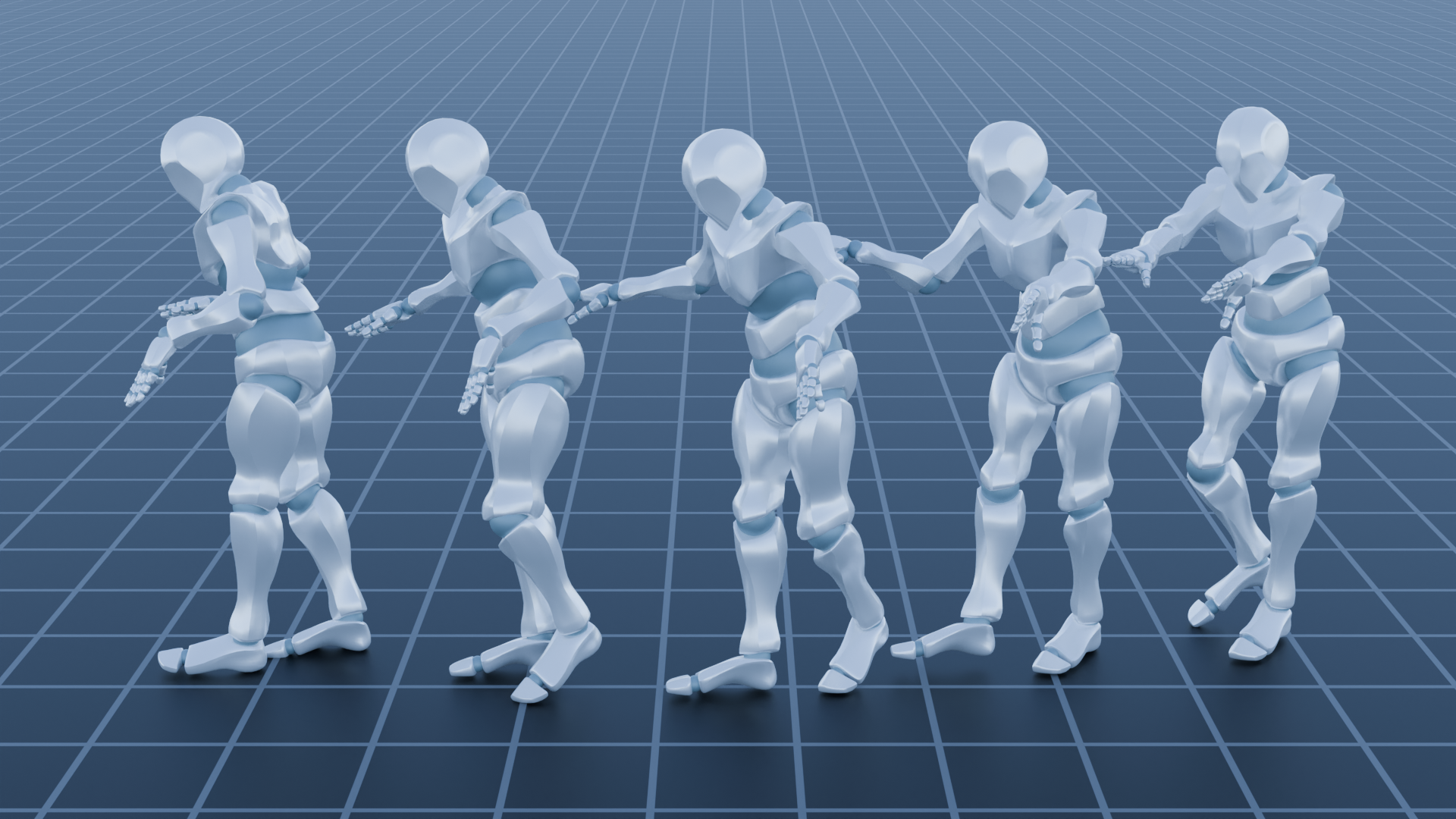}
        \caption{}
        \label{fig:change A b}
    \end{subfigure}
    \hspace{20px} 
    \begin{subfigure}{0.275\textwidth}
        \centering
        \includegraphics[width=\linewidth]{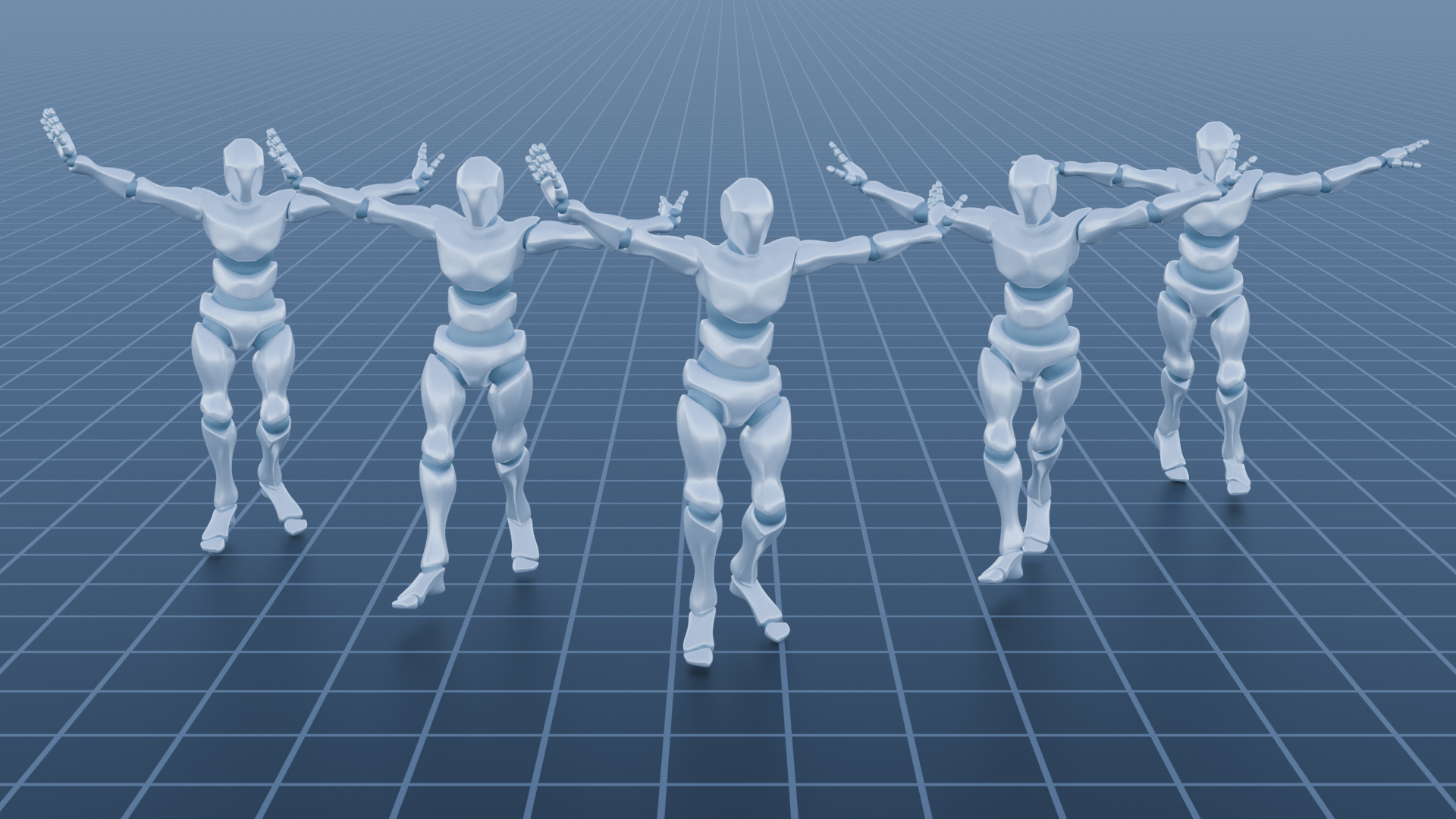}
        \caption{}
        \label{fig:change A c}
    \end{subfigure}


    \begin{subfigure}{0.275\textwidth}
        \centering
        \includegraphics[width=\linewidth]{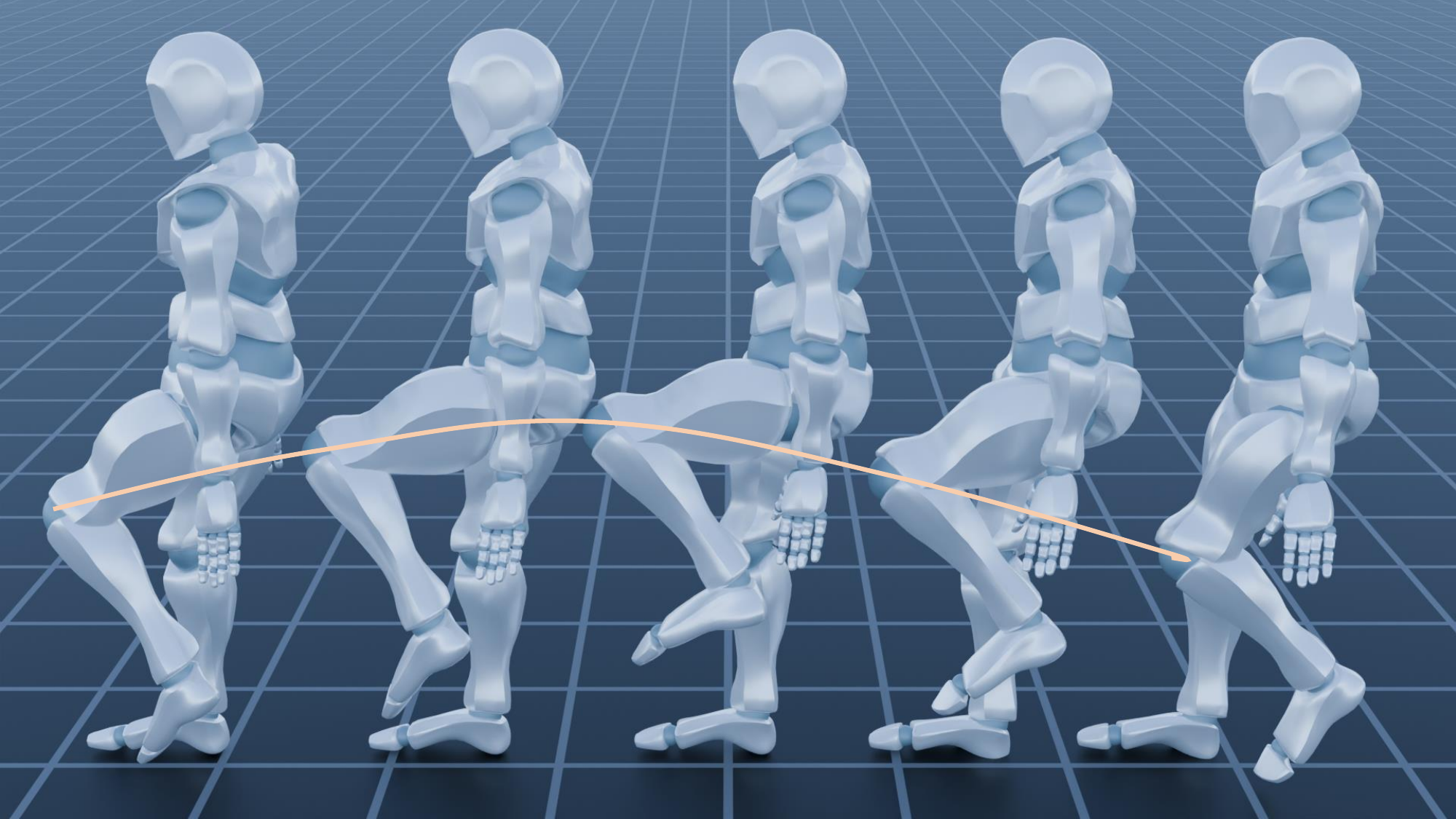}
        \caption{}
        \label{fig:change A d}
    \end{subfigure}%
    \hspace{20px} 
    \begin{subfigure}{0.275\textwidth}
        \centering
        \includegraphics[width=\linewidth]{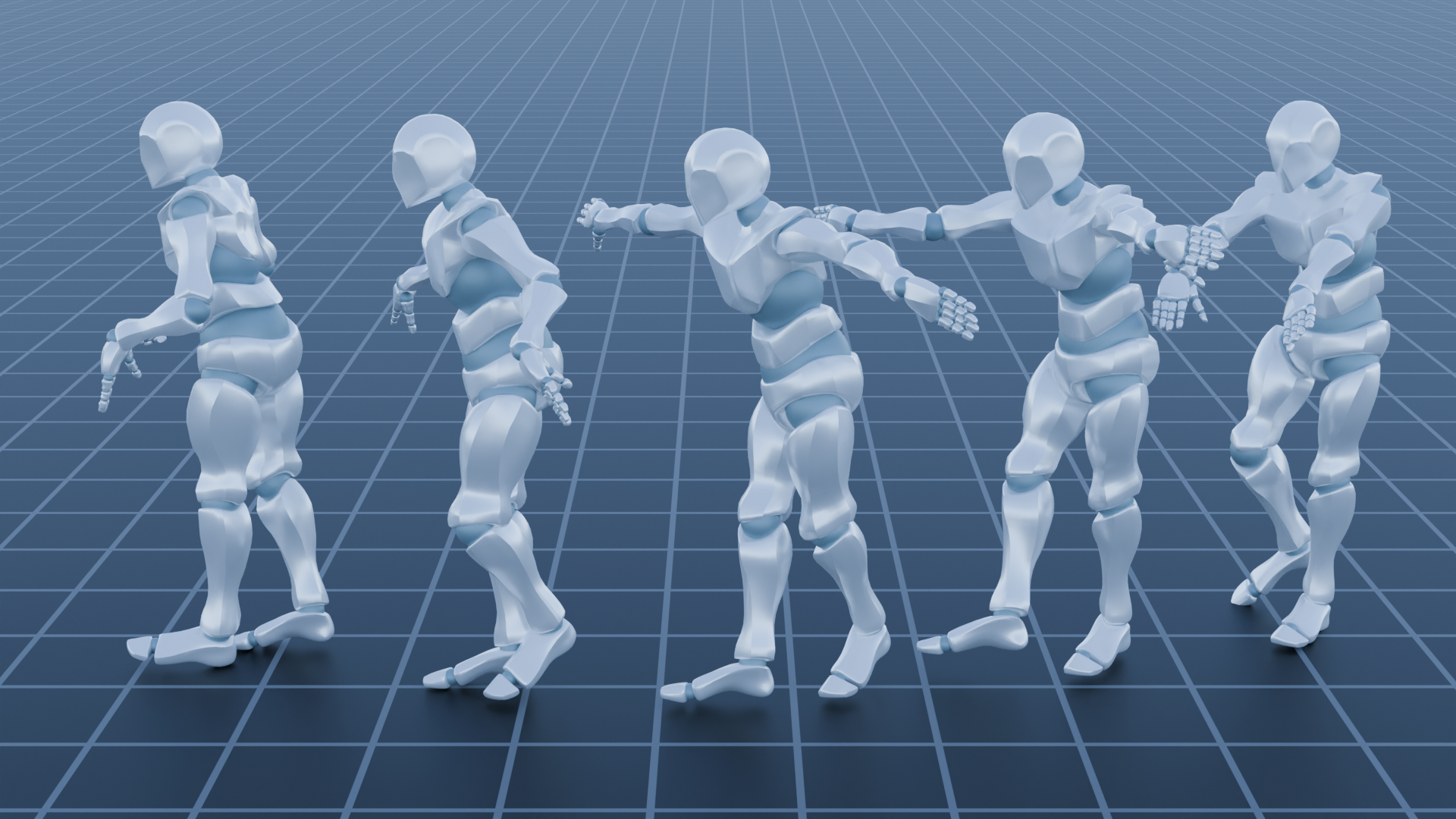}
        \caption{}
        \label{fig:change A e}
    \end{subfigure}
    \hspace{20px} 
    \begin{subfigure}{0.275\textwidth}
        \centering
        \includegraphics[width=\linewidth]{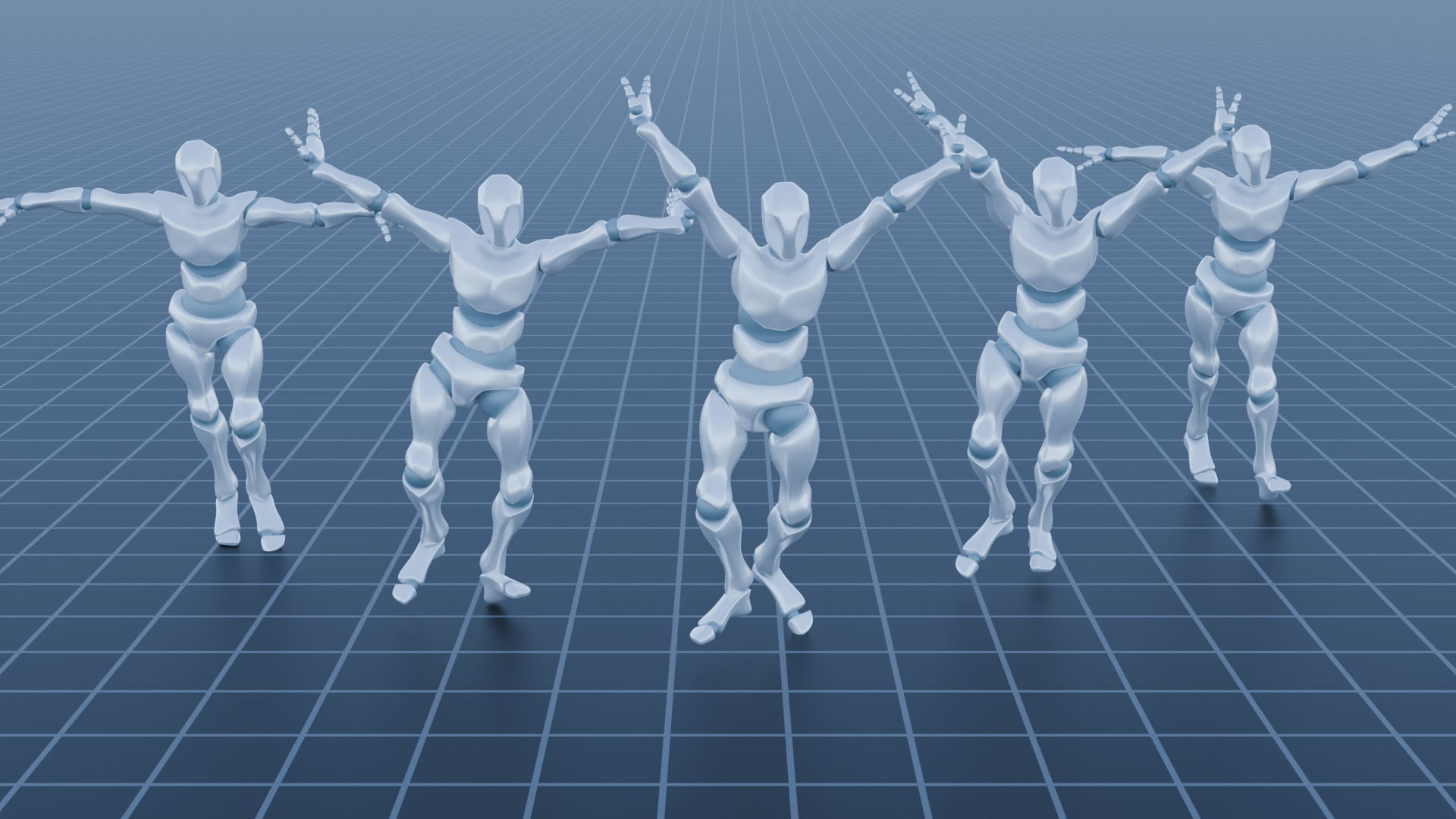}
        \caption{}
        \label{fig:change A f}
    \end{subfigure}

    \begin{subfigure}{0.325\textwidth}
        \centering
        \includegraphics[width=\linewidth]{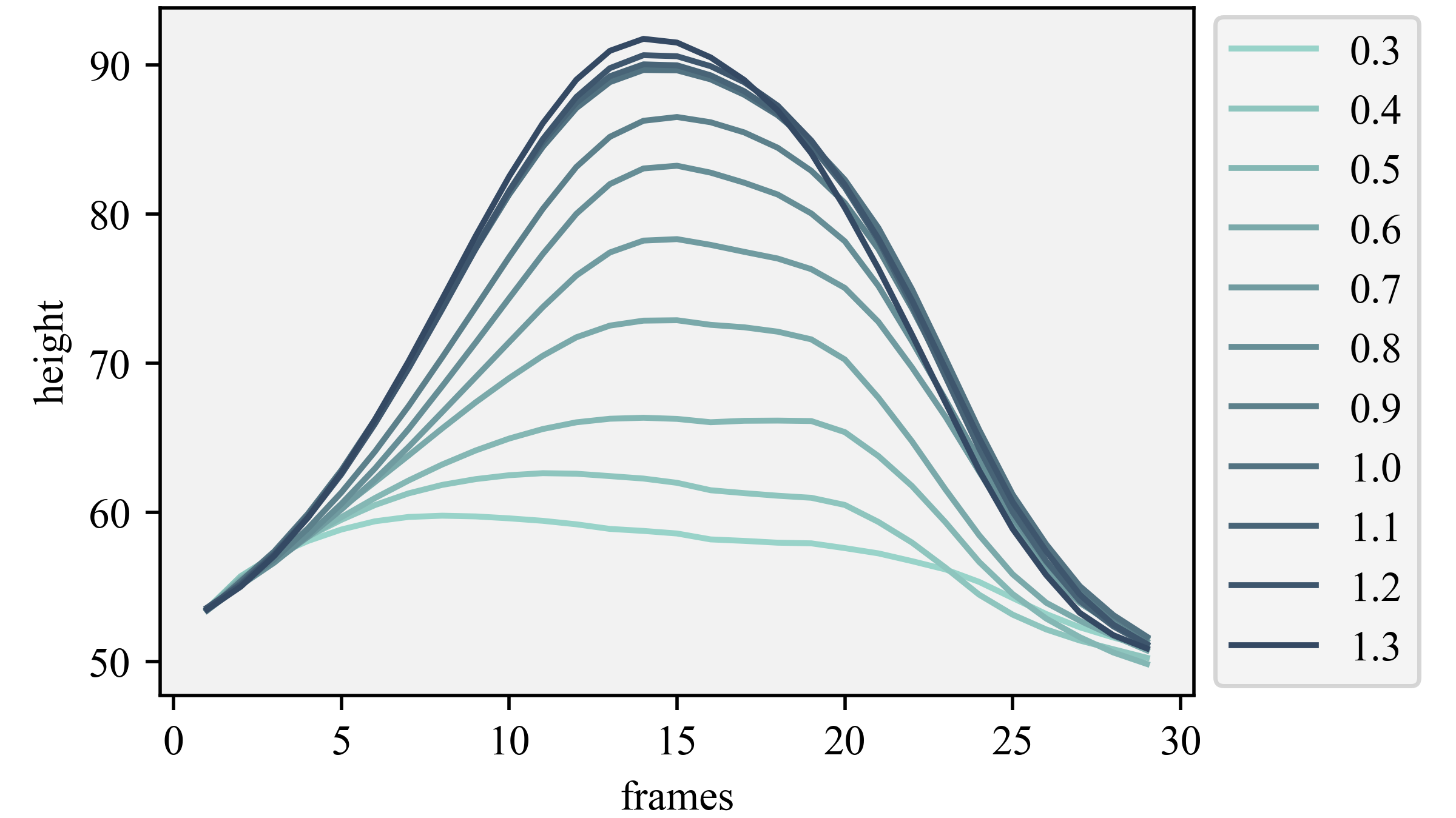}
        \caption{}
        \label{fig:change A g}
    \end{subfigure}%
    \hspace{0px} 
    \begin{subfigure}{0.325\textwidth}
        \centering
        \includegraphics[width=\linewidth]{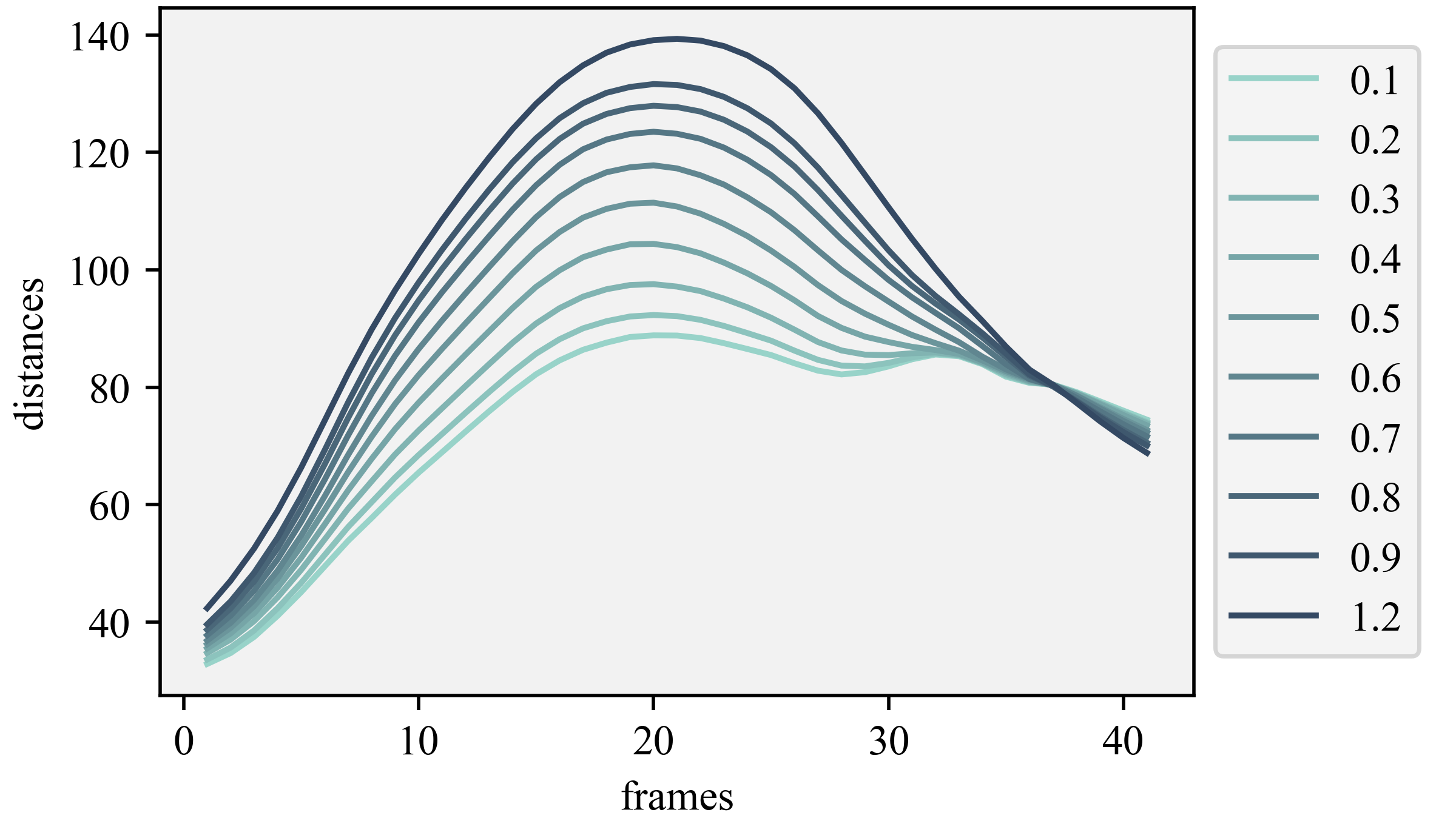}
        \caption{}
        \label{fig:change A h}
    \end{subfigure}
    \hspace{0px} 
    \begin{subfigure}{0.325\textwidth}
        \centering
        \includegraphics[width=\linewidth]{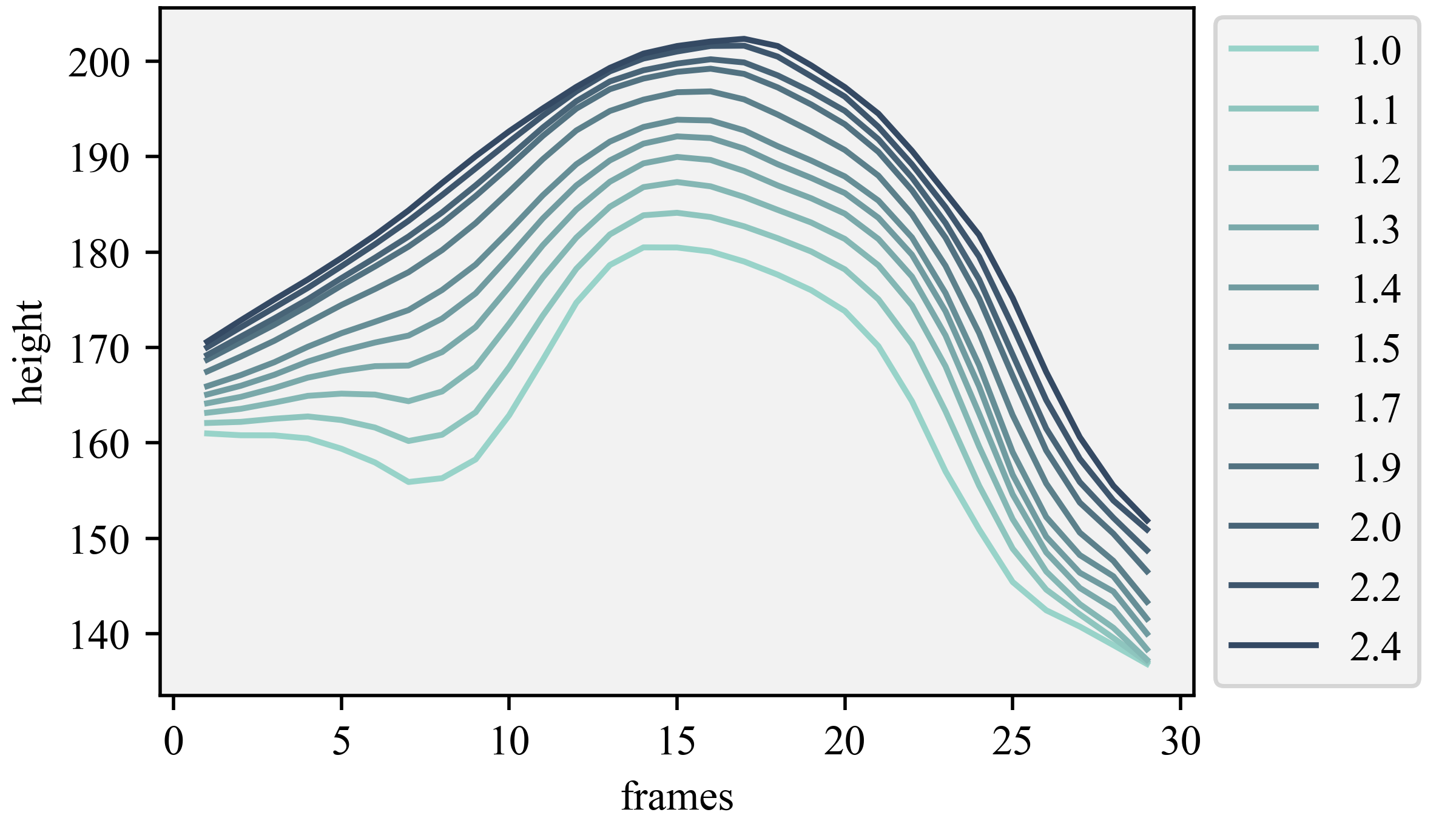}
        \caption{}
        \label{fig:change A i}
    \end{subfigure}
    \vspace{-5pt}
    \caption{Results of adjusting the amplitude of body part phases}
    \label{fig:change A}
\end{figure*}
\subsection{Experimental Settings}
\paragraph{Body part grouping.}

Human motion possesses inherent structural characteristics due to inter-joint coordination. We implement two physiologically-grounded body part divisions (Fig.\ref{fig:body part}): (1) The bipartite decomposition follows biomechanical principles from~\cite{lu2024mobile,ji2024exbody2}, separating upper and lower extremities to preserve limb-specific dynamics; (2) The five-segment scheme builds upon kinematic chain analyses in ~\cite{agrawal2024skel,li2024walkthedog,wu2023analysis,chavez2024asv}, isolating torso, bilateral arms and legs to capture finer motion semantics. 
This explicit part-wise decomposition facilitates the decoupling of movement patterns while preserving physiological coordination, providing a significant advantage over automated skeleton-aware convolutions~\cite{agrawal2024skel} for phase-based motion editing.
\vspace{-8pt}
\paragraph{Dataset.}
Following~\cite{tang2023RSMT}, we utilize the 100STYLE motion dataset~\cite{mason2022real}. We remove all wrist and thumb joints to retarget the motion to a skeleton with 23 joints. Additionally, we downsample the motion sequences to 30 fps and augment them using mirroring and random time cropping~\cite{jang2022motion}. We clip the motion manifold into segments of 120 frames, with each segment overlapping by 20 frames. The motion orientation is adjusted so that the first frame faces the X-axis~\cite{holden2016deep}. We employ global joint directions rotating along a forward and upward vector \(r \in \mathbb{R}^6\)~\cite{zhang2018mode}. The motion \(\textbf{M} \in \mathbb{R}^{23 \times 12 \times 120}\) encompasses global joint positions (\(\mathbb{R}^3\)), velocities (\(\mathbb{R}^3\)), and rotations (\(\mathbb{R}^6\)).
\vspace{-8pt}
\paragraph{Metrics.}
We evaluate the movements in terms of reconstruction accuracy and foot skating artifacts. The reconstruction accuracy is calculated using the averaged L2 distance of global joint positions and the Normalized Power Spectrum Similarity (NPSS) in the joint angle space~\cite{harvey2020robust}. The foot skating artifacts~\cite{zhang2018mode} are quantified based on the averaged foot velocity \(v_f\) when the foot height \(h\) is below a threshold \(H = 2.5\): \(L_f = v_f \cdot \text{clamp}(2 - 2^{h/H}, 0, 2).\)
\vspace{-8pt}
\paragraph{Implementation Details.}
Our \textbf{network architecture} is configured as follows: The Body Part Mixture of Experts (BPMoE) employs 8 parallel expert networks. All encoder modules (excluding the style encoder) follow an identical structure: a feed-forward network comprising one hidden layer of 512 neurons and an output layer of 256 neurons, with PLU activation applied after each linear transformation. The temporal modeling component utilizes an LSTM with 1024 hidden units for motion sequence processing. The phase decoder architecture consists of two successive feed-forward layers – the first with 512 units and the second with 256 units – both employing Exponential Linear Unit (ELU) activation functions.

To accelerate training convergence, we randomly select a 25-frame window from the 120-frame clips for each step instead of using the full sequence. We employ AMSGrad with parameters \((\beta_1 = 0.5, \beta_2 = 0.9)\) and a learning rate of \(1e - 3\). For training the sampler, we shuffle all style sequences randomly. To ensure model robustness across varying time lengths, we randomly sample a sequence starting at a length of 20 frames and linearly increase it to 40 frames at each epoch, similar to~\cite{harvey2020robust}. We use the same AMSGrad setup as before, with a weight decay of \(1e - 4\) for training the style encoder to prevent overfitting, while the weight decay for other modules is set to 0. After training the BPMoE, we remove its encoder, fix its decoder, and connect the motion sampler to the fixed decoder to train the sampler. This training process typically takes around a day. We conduct our experiments using an NVIDIA 4090 24G GPU. Our method takes, on average, 1.87 ms to synthesize one frame, which is sufficient for real-time applications.
\subsection{Comparison with State-of-the-Arts}
\paragraph{Comparison under Seen Styles.}
We evaluate our framework against CVAE~\cite{tang2022CVAE}, RSMT~\cite{tang2023RSMT} and PhaseMIB~\cite{starke2023motion} on the 100STYLE dataset, with results in Table \ref{tab:Comparison A}. For short sequences, RSMT~\cite{tang2023RSMT} and CVAE~\cite{tang2022CVAE} exhibit better controllability due to their interpolation-focused design that minimizes temporal feature complexity. However, this simplification becomes detrimental for longer sequences, where CVAE~\cite{tang2022CVAE}'s insufficient temporal modeling and RSMT~\cite{tang2023RSMT}'s ineffective temporal encoding lead to performance degradation.
While PhaseMIB~\cite{starke2023motion} improves long-sequence generation through bidirectional control, its whole-body phase representation intrinsically couples limb dynamics, causing style entanglement that limits part-wise motion control, which is a fundamental limitation shared with prior approaches.

Our approach decouples full-body movements and captures temporal information that better aligns with human structural characteristics. Therefore, when confronted with tasks requiring the generation of long sequences, we excel in controllability and motion quality. Specifically, dividing the body into two parts accounts for the collaborative effects of the upper and lower body separately, leading to a better representation of overall frequency domain information and resulting in superior performance on the NPSS metric. Dividing the body into five parts further captures the intrinsic periodicity of the limbs, enhancing the frequency domain representation for individual limbs and thereby outperforming the skating metric. As both models excel in representing temporal information, they surpass baseline in terms of the averaged L2 distance of global joint positions.
\vspace{-8pt}
\paragraph{Comparison under Unseen Styles.}
\label{sec:Generating styles with body part changes}
We evaluate our method’s generalization to unseen styles caused by body part modifications. Three test sets are prepared, involving changes to the upper limbs (mirrored in the lower limbs), the left upper limb, and the right lower limb. A major challenge in localized modifications is maintaining kinematic consistency. Direct stitching (see supplementary materials) often introduces artifacts such as simultaneous hand and foot movements. To avoid inconsistent hip-limb coordination, we introduce upper-body influence on lower-body dynamics for automatic inter-joint conflict correction. By replacing the rotations and positions of selected body parts with those from the target style, we achieve localized modifications.
For each test sequence, a random target style is applied. RSMT~\cite{tang2023RSMT} encodes style from full-body motion sequences via a convolutional neural network, for which we provide both original and modified inputs. In contrast, our method replaces the selected body part’s phase with that of the target style. 

As shown in Table \ref{tab:compare in AB}, our phase-aware replacement outperforms direct stitching in preserving limb coordination (see supplementary video). CVAE~\cite{tang2022CVAE}, without explicit style control, predicts the most probable motion. The similarity between RSMT (original) and RSMT (modified) suggests motion-sequence-based encoding has limited impact. PhaseMIB~\cite{starke2023motion} struggles with body-part variations due to its teacher-forcing paradigm, which limits extrapolation beyond training data. In contrast, our part-wise phase representation and autoregressive generation enable dynamic inter-limb coordination during synthesis.
\subsection{Ablation Study on Style Encoder.}
We evaluate the effectiveness of the body-part-phase-based style encoder in our framework, as shown in Table \ref{tab:style A}. In this study, we replace the style encoder with convolutional neural networks that utilize the full-body motion sequence. The construction of the test data is consistent with that described in Section \ref{sec:Generating styles with body part changes}. We also evaluate the effectiveness of the style encoder under conditions of body part style variation. When style encoding relies on motion sequences, it fails to effectively capture style information, leading to inferior motion quality compared to our framework. The results of the ablation experiments highlight the significance of style encoding based on body part phases, emphasizing its contribution to the effectiveness and quality of our framework.
\begin{figure*}[thbp]
    \centering
    \begin{subfigure}{0.275\textwidth}
        \centering
        \includegraphics[width=\linewidth]{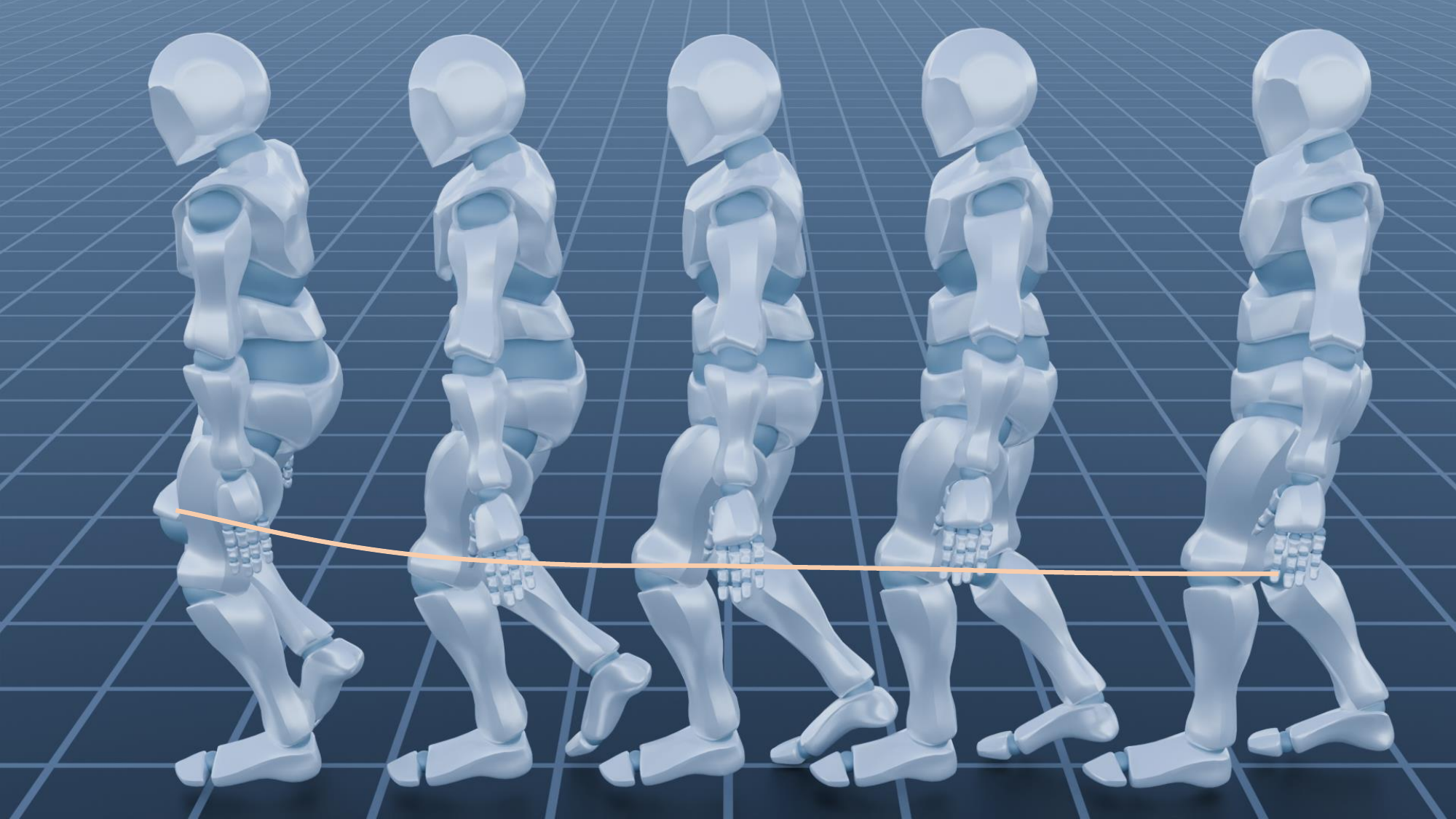}
        \caption{}
        \label{fig:change S a}
    \end{subfigure}%
    \hspace{20px} 
    \begin{subfigure}{0.275\textwidth}
        \centering
        \includegraphics[width=\linewidth]{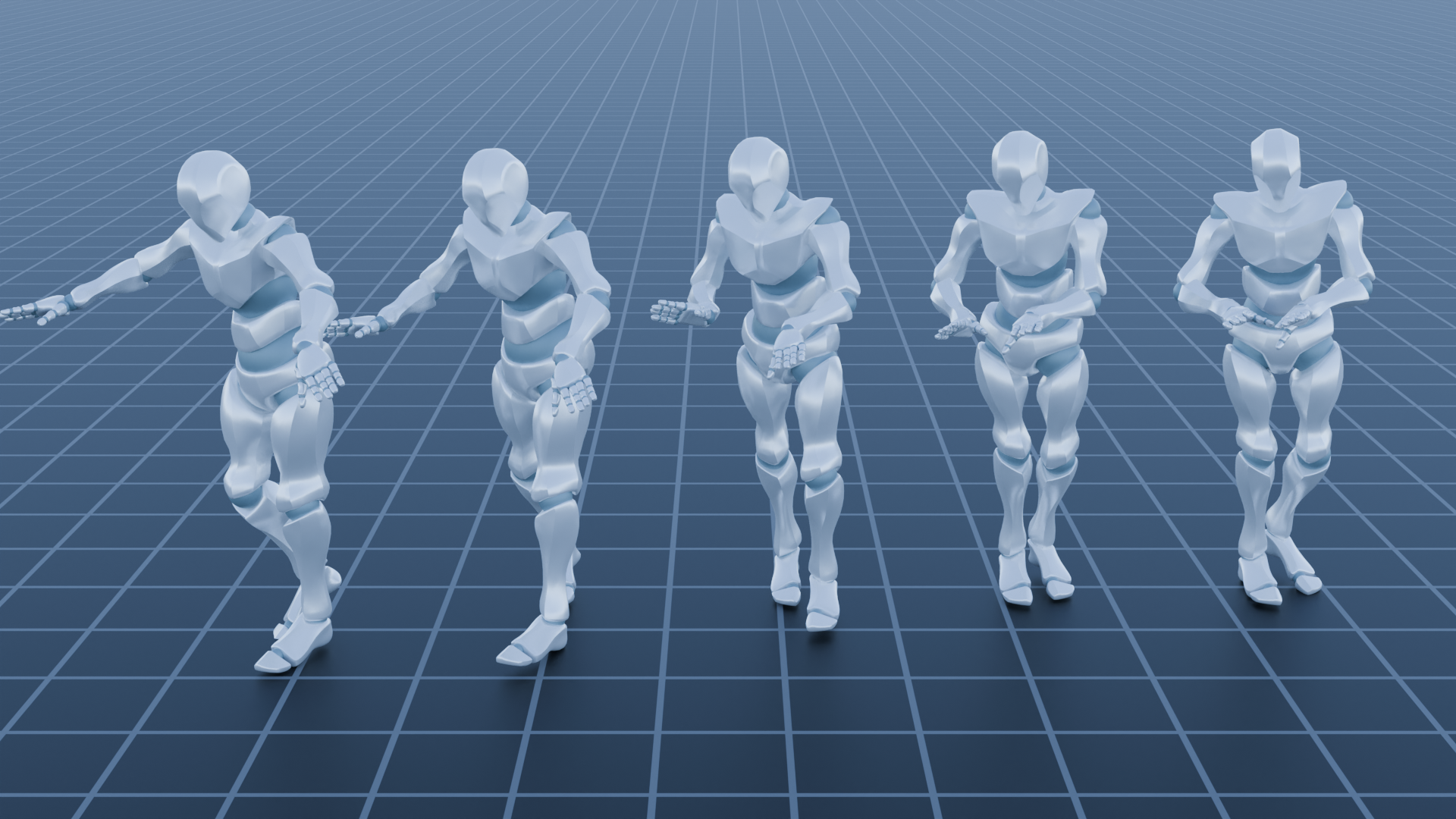}
        \caption{}
        \label{fig:change S b}
    \end{subfigure}
    \hspace{20px} 
    \begin{subfigure}{0.275\textwidth}
        \centering
        \includegraphics[width=\linewidth]{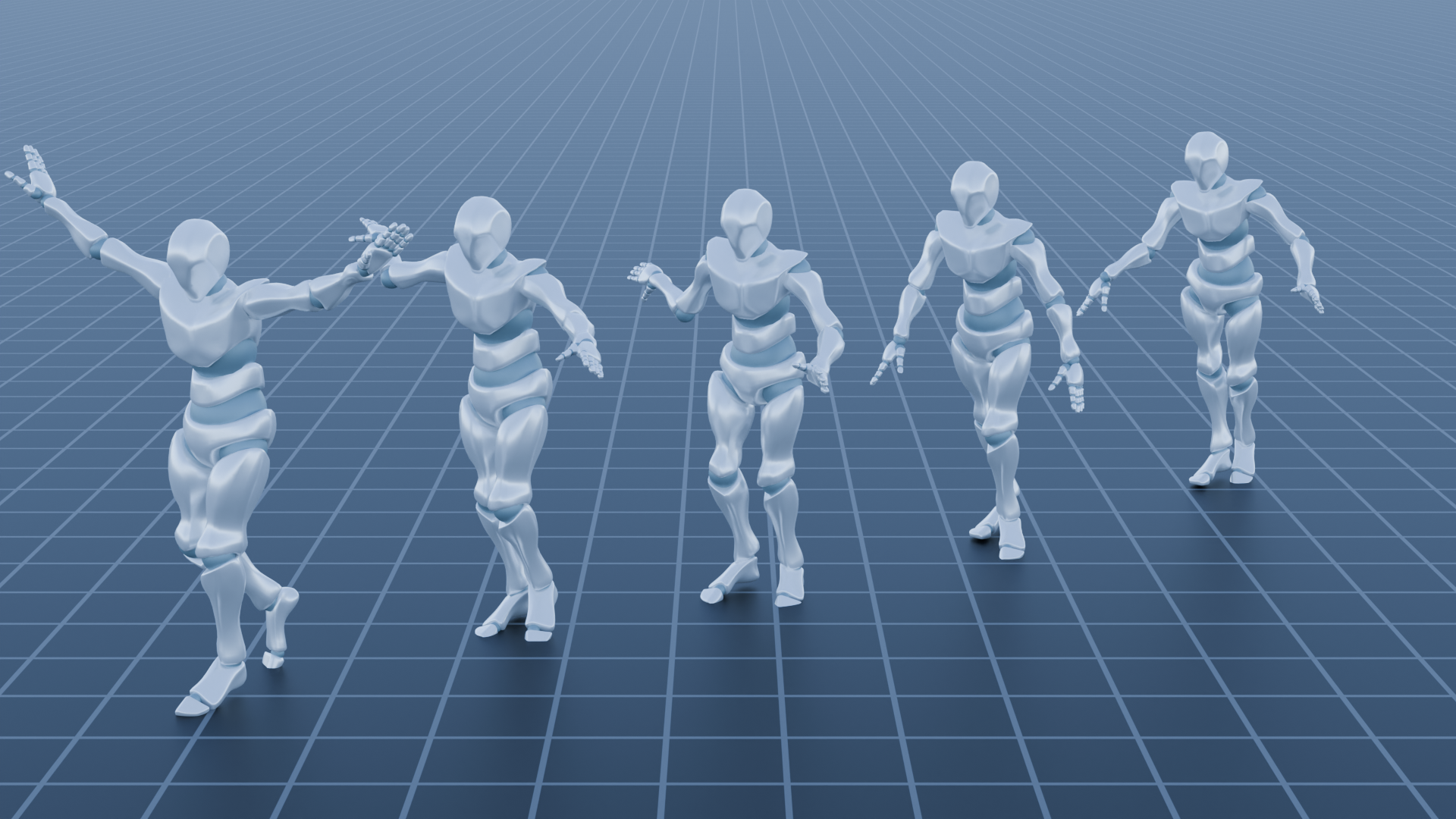}
        \caption{}
        \label{fig:change S c}
    \end{subfigure}


        \begin{subfigure}{0.275\textwidth}
        \centering
        \includegraphics[width=\linewidth]{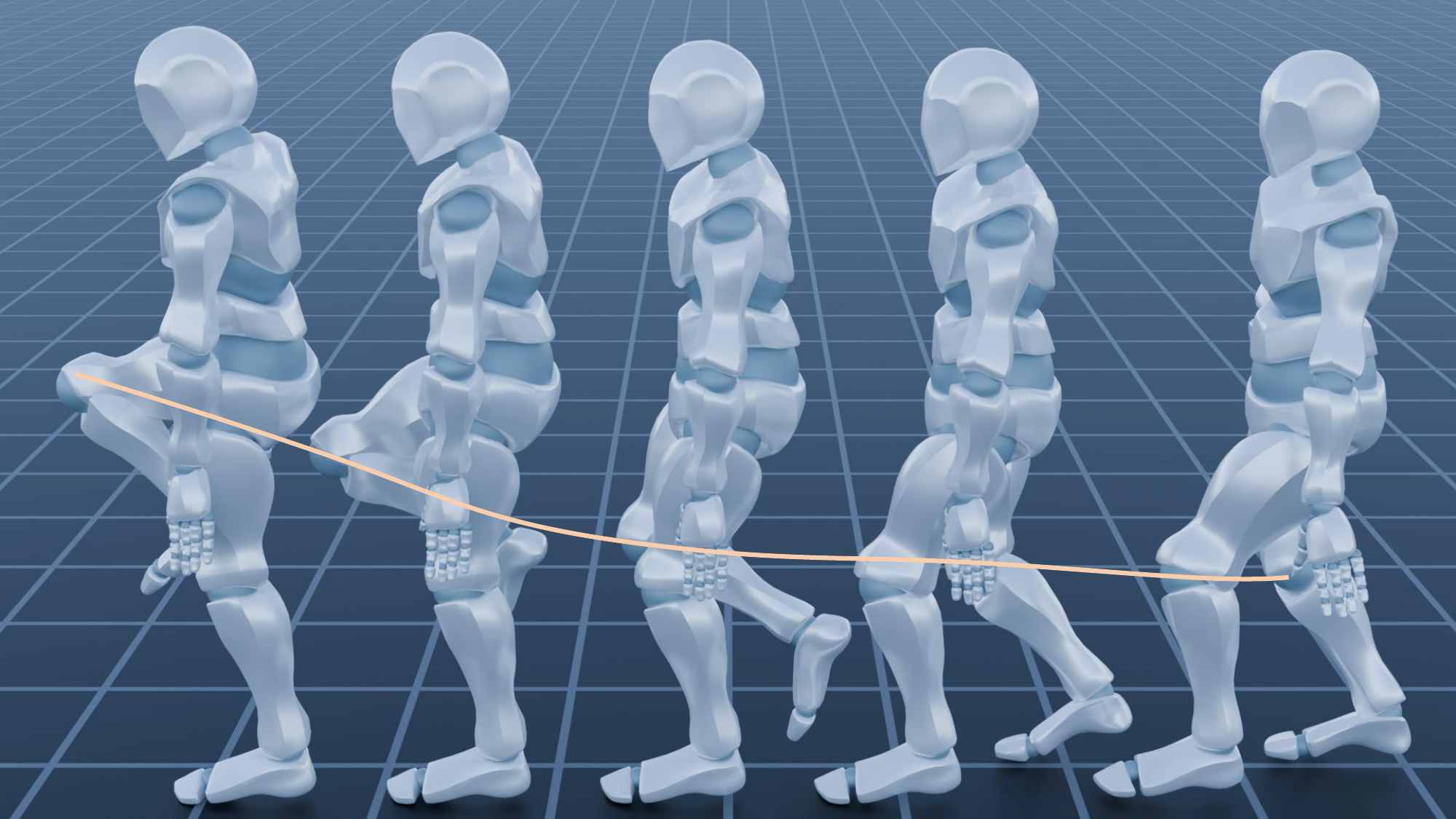}
        \caption{}
        \label{fig:change S d}
    \end{subfigure}%
    \hspace{20px} 
    \begin{subfigure}{0.275\textwidth}
        \centering
        \includegraphics[width=\linewidth]{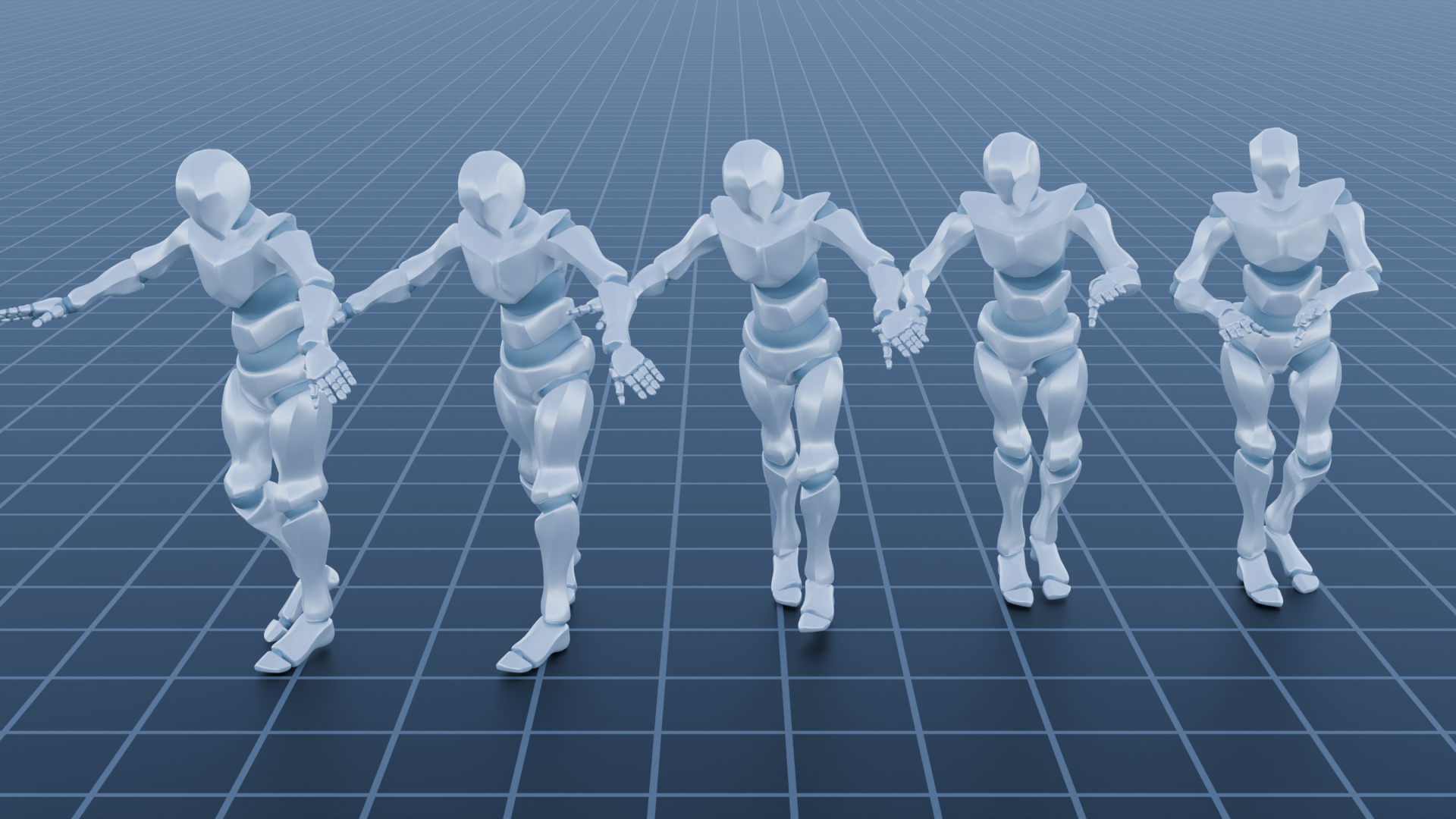}
        \caption{}
        \label{fig:change S e}
    \end{subfigure}
    \hspace{20px} 
    \begin{subfigure}{0.275\textwidth}
        \centering
        \includegraphics[width=\linewidth]{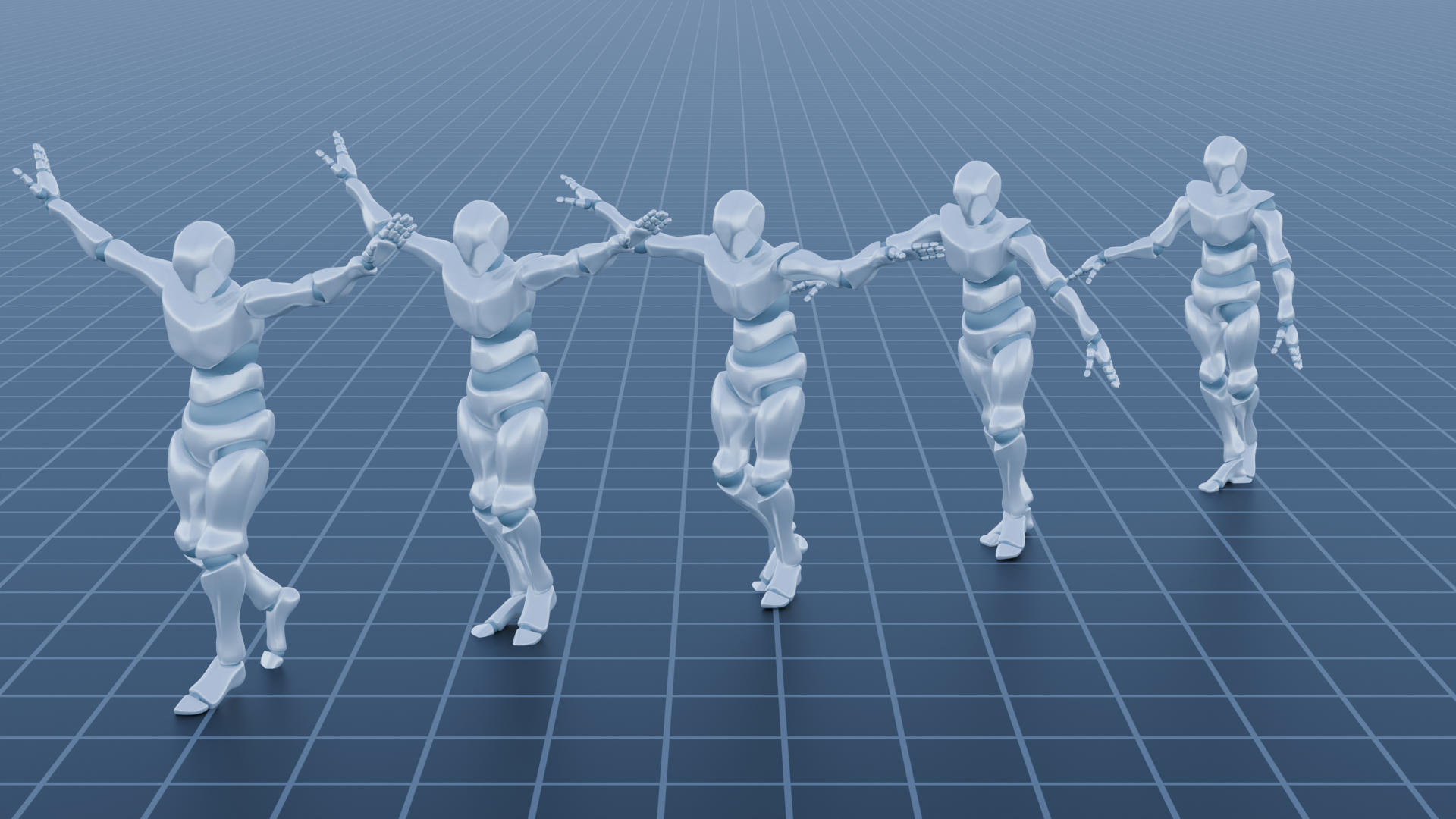}
        \caption{}
        \label{fig:change S f}
    \end{subfigure}

    \begin{subfigure}{0.31\textwidth}
        \centering
        \includegraphics[width=\linewidth]{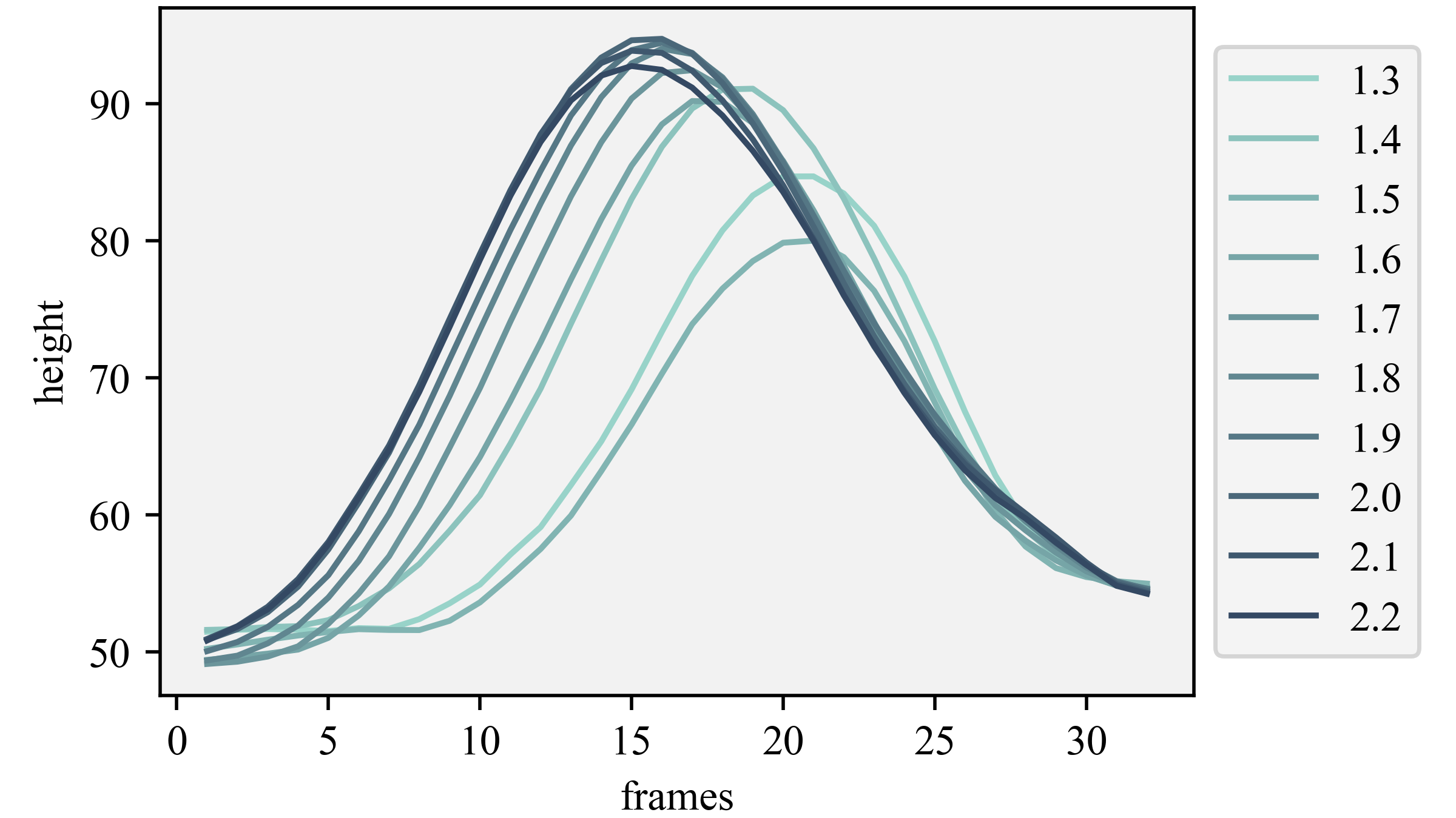}
        \caption{}
        \label{fig:change S g}
    \end{subfigure}%
    \hspace{0px} 
    \begin{subfigure}{0.31\textwidth}
        \centering
        \includegraphics[width=\linewidth]{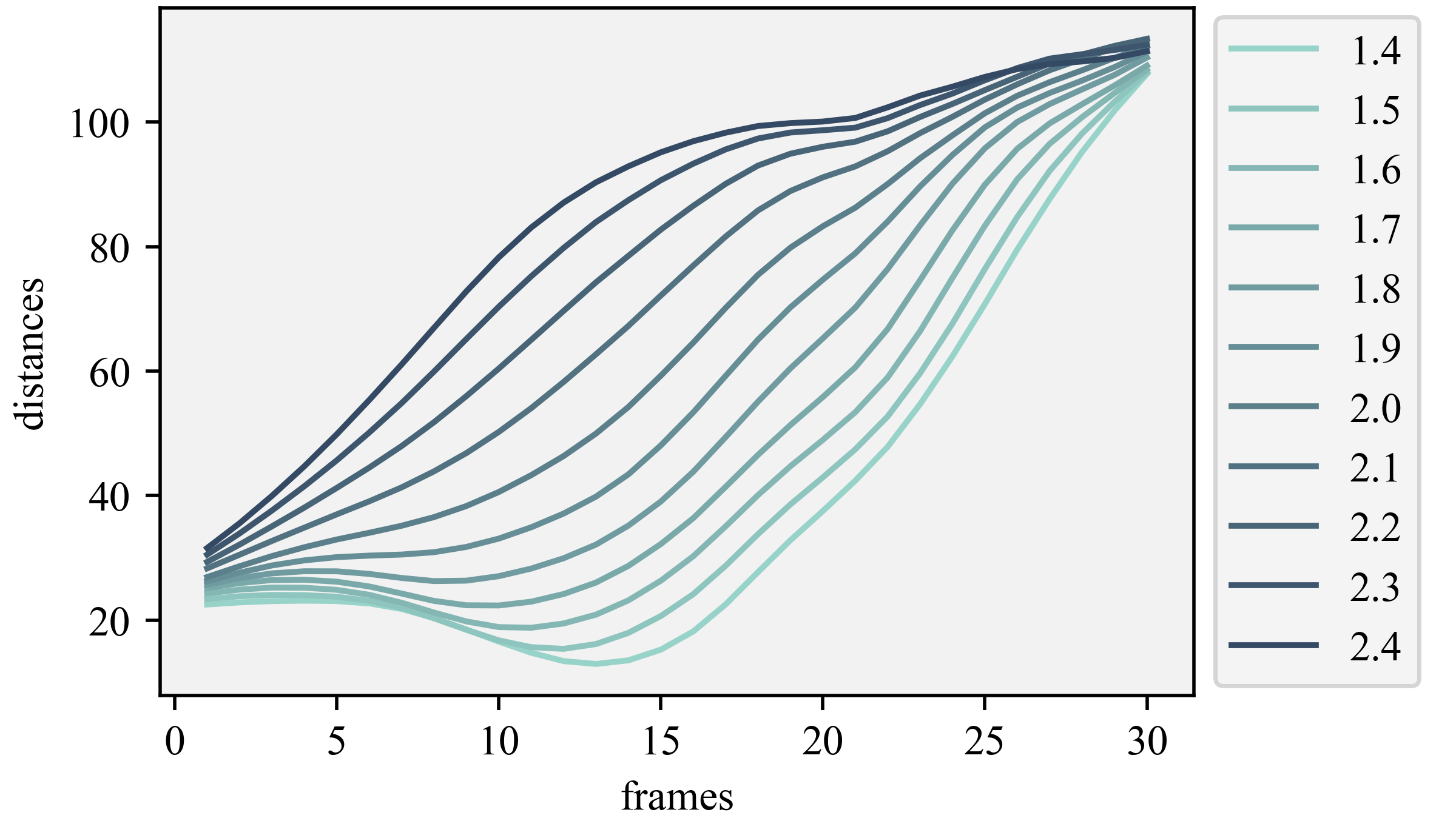}
        \caption{}
        \label{fig:change S h}
    \end{subfigure}
    \hspace{0px} 
    \begin{subfigure}{0.31\textwidth}
        \centering
        \includegraphics[width=\linewidth]{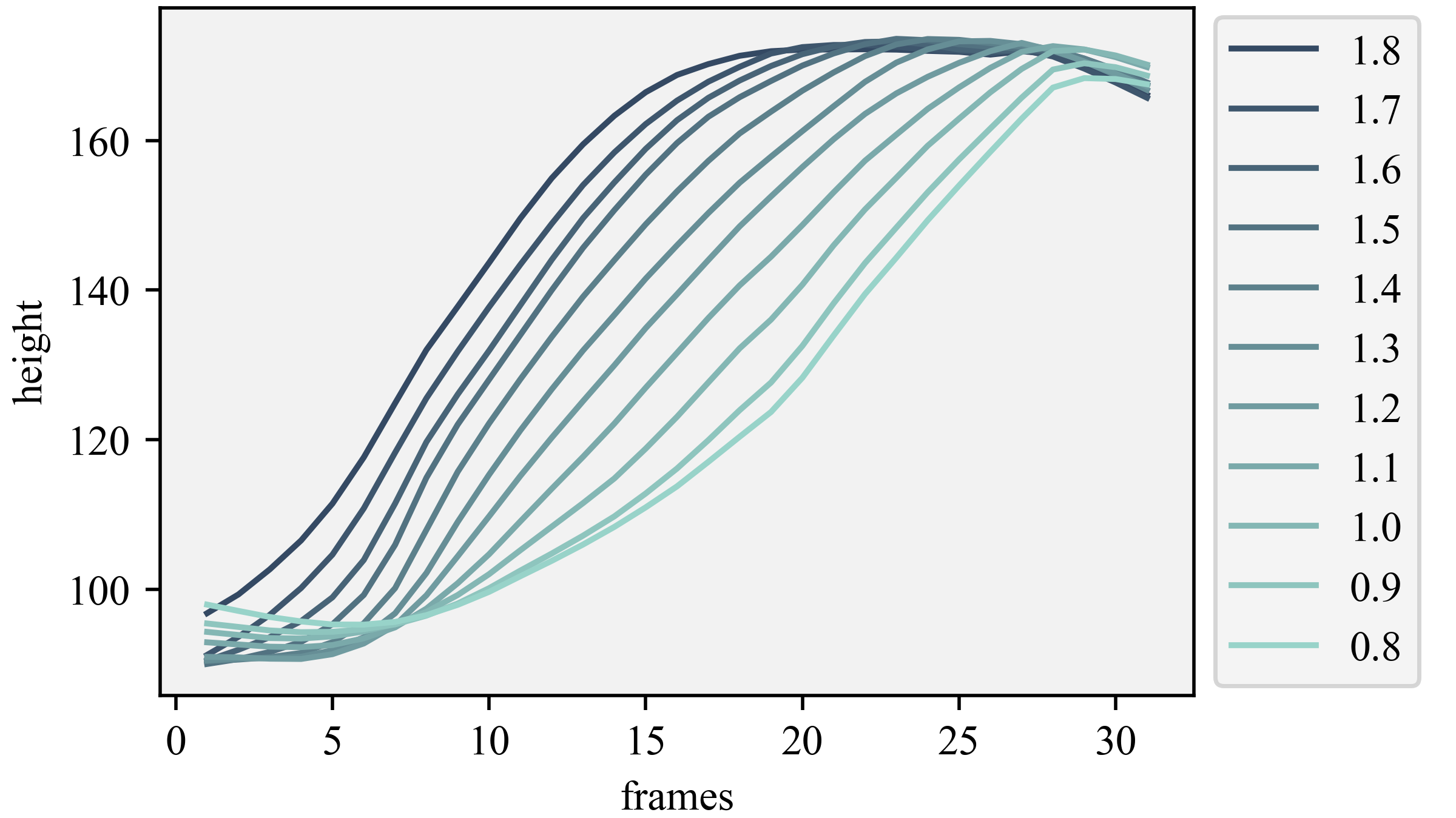}
        \caption{}
        \label{fig:change S i}
    \end{subfigure}
    
    \vspace{-5pt}
    \caption{Results of adjusting the frequency of body part phases}
    \label{fig:change S}
\end{figure*}
\subsection{Body Part Motion Adjustment}
We scale the body part phases \(A\) and \(S\) by a factor and use the adjusted values to compute the phase per formula \ref{eq:phase}. The scaled phase serves as an input to the sampler, predicting the next frame’s phase and updating gating network parameters. To maintain coordination, unselected body parts may exhibit slight variations.

We evaluate three styles: \textbf{\textit{HighKnees}}, \textbf{\textit{Swimming}}, and \textbf{\textit{Flapping}}. For each, representative motion segments are selected, with frames \(i\) and \(j\) marking the start and end. The first two rows of each column in the figures show motion snapshots at five sampled points, while the third row illustrates changes in key metrics from \(i\) to \(j\) (in Figure \ref{fig:change S i}, the ending frame extends 17 frames beyond \(j\)).
\paragraph{Amplitude Control.}
As shown in Figure \ref{fig:change A}, our framework adjusts motion amplitude by modifying phase amplitudes. For \textbf{\textit{HighKnees}}, reducing lower limb phase amplitudes to 50\% of their original values (Figure \ref{fig:change A a}) decreases knee height, compared to the unmodified version (Figure \ref{fig:change A d}). For \textbf{\textit{Swimming}}, reducing upper limb phase amplitudes to 50\% (Figure \ref{fig:change A b}) narrows the wrist distance, while Figure \ref{fig:change A e} shows the original. For \textbf{\textit{Flapping}}, Figure \ref{fig:change A f} increases upper limb phase amplitudes to 250\%, raising hand height compared to Figure \ref{fig:change A c}. Figures \ref{fig:change A g}–\ref{fig:change A i} show the effect of different scaling factors on leg lift amplitude, hand distance, and hand height.
\vspace{-8pt}
\paragraph{Frequency Control.}
Figure \ref{fig:change S} illustrates the impact of modifying phase frequencies on motion speed. 
For the \textbf{\textit{HighKnees}} style, increasing the lower limb phase frequency to 250\% of its original value (Figure \ref{fig:change S d}) results in the knee reaching its peak height significantly faster compared to the unmodified motion (Figure \ref{fig:change S a}). This effect is quantitatively shown in Figure \ref{fig:change S g}, where the leg lift amplitude at different scaling factors is plotted over time. A higher frequency accelerates movement, while slower frequencies cause the knee to reach a lower peak before descending to maintain consistency with the final keyframe.
For the \textbf{\textit{Swimming}} style, doubling the phase frequency of the upper limbs (Figure \ref{fig:change S e}) speeds up arm movement, making them reach target positions more quickly than in the unmodified version (Figure \ref{fig:change S b}). Figure \ref{fig:change S h} further illustrates how the distance between the hands varies under different frequency scaling factors, demonstrating that increased frequency leads to faster hand convergence.
For the \textbf{\textit{Flapping}} style, reducing the upper limb phase frequency to 50\% of its original value (Figure \ref{fig:change S c}) slows down the arm motion, while Figure \ref{fig:change S f} shows the original unmodified pace. Despite these changes in motion speed, Figure \ref{fig:change S i} confirms that the amplitude remains consistent, as both the highest and lowest points of hand movement remain unaffected by frequency variations.

\section{Conclusion}
In this paper, we introduce a novel framework for stylized motion in-betweening with a focus on body-part-level control. Leveraging phase-related techniques and periodic autoencoders, our approach improves both the diversity and controllability of animated sequences. Experimental results demonstrate that our method outperforms existing approaches in generating realistic animations, enabling precise adjustments to individual limb movements while preserving overall motion coherence. These advancements establish our framework as a valuable tool for expressive motion synthesis, with applications in gaming, virtual reality, and beyond.

{
    \small
    \bibliographystyle{ieeenat_fullname}
    \bibliography{main}
}

\end{document}